%% file: ms.tex
\documentclass{article}
\usepackage[margin=1in]{geometry}
\usepackage{fullpage}
\usepackage[numbers]{natbib}
\usepackage[utf8]{inputenc} 
\usepackage[T1]{fontenc}    
\usepackage{hyperref}       
\usepackage{url}            
\usepackage{booktabs}       
\usepackage{amsfonts}       
\usepackage{amsthm}
\usepackage{nicefrac}       
\usepackage{microtype}      
\usepackage{subcaption}
\usepackage{graphicx}
\usepackage{algorithm}
\usepackage{algpseudocode}

\usepackage{personalmacros}
\usepackage{articleheader}

\input{mathmacros.tex}

\input{AMS_commands.tex}

\input{variables.tex}
\newcommand{\xbar}{\bar \xx}

\title{Minimax Estimation of Bandable Precision
  Matrices\footnote{Conference version to be presented at NIPS 2017, Long Beach, CA}}

\author{
  Addison Hu\footnote{
      Addison graduated from Yale in May 2017.  Up-to-date contact information
      may be found at \url{http://huisaddison.com/}.
  }  \qquad\qquad Sahand Negahban  \\\\
  Department of Statistics and Data Science\\
  Yale University\\
  New Haven, CT 06520 \\
  \texttt{\{addison.hu, sahand.negahban\}@yale.edu} \\
}

\begin{document}

\maketitle

\begin{abstract}
    The inverse covariance matrix provides considerable insight for
    understanding statistical models in the multivariate setting.  In
    particular, when the distribution over variables is assumed to be
    multivariate normal, the sparsity pattern in the inverse covariance matrix,
    commonly referred to as the precision matrix, corresponds to the adjacency
    matrix representation of the Gauss-Markov graph, which encodes conditional
    independence statements between variables.  Minimax results under the
    spectral norm have previously been established for covariance matrices,
    both sparse and banded, and for sparse precision matrices.  We establish
    minimax estimation bounds for estimating banded precision matrices under
    the spectral norm. Our results greatly improve upon the existing
    bounds; in particular, we find that the minimax rate for estimating banded
    precision matrices matches that of estimating banded covariance matrices.
    The key insight in our analysis is that we are able to obtain barely-noisy
    estimates of \(k \times k\) subblocks of the precision matrix by inverting
    slightly wider blocks of the empirical covariance matrix along the
    diagonal.  Our theoretical results are complemented by experiments
    demonstrating the sharpness of our bounds.
\end{abstract}

\section{Introduction}\label{sect:intro}
\input{intro.tex}

\section{Background and problem set-up}\label{sect:methodology}
\input{methodology.tex}

\section{Rate optimality under the spectral norm}\label{sect:optimality}
\input{optimality.tex}

\section{Experimental results}\label{sect:simulations}
\input{simulations.tex}

\section{Discussion}
\label{sec:discussion}
\input{discussion.tex}

\subsubsection*{Acknowledgements}
We would like to thank Harry Zhou for
stimulating discussions regarding matrix estimation problems. SN
acknowledges funding from NSF Grant DMS 1723128.

\newpage
{\footnotesize{
\bibliography{final}{}
\bibliographystyle{plain}}}
\newpage
\appendix
\input{appendix.tex}

\end{document}

%% file: mathmacros.tex
\DeclareMathOperator{\Card}{Card}

\newlength{\widebarargwidth}
\newlength{\widebarargheight}
\newlength{\widebarargdepth}

%% file: AMS_commands.tex
\theoremstyle{plain}

\newtheorem{theo}{Theorem}[section]

\newtheorem{lem}{Lemma}[section]
\newtheorem{prop}{Proposition}[section]
\newtheorem{cor}{Corollary}[section]

\theoremstyle{definition} 

\newtheorem{nota}{Notation}[section]
\newtheorem{de}{Definition}[section]
\newtheorem{exa}{Example}[section]
\newtheorem{as}{Assumption}[section]
\newtheorem{alg}{Algorithm}[section]

\newcommand{\btheo}{\begin{theo}}
\newcommand{\bde}{\begin{de}}
\newcommand{\ble}{\begin{lem}}
\newcommand{\bpr}{\begin{prop}}
\newcommand{\bno}{\begin{nota}}
\newcommand{\bex}{\begin{exa}}
\newcommand{\bcor}{\begin{cor}}
\newcommand{\spro}{\begin{proof}}
\newcommand{\bas}{\begin{as}}
\newcommand{\balg}{\begin{alg}}

\newcommand{\etheo}{\end{theo}}
\newcommand{\ede}{\end{de}}
\newcommand{\ele}{\end{lem}}
\newcommand{\epr}{\end{prop}}
\newcommand{\eno}{\end{nota}}
\newcommand{\eex}{\end{exa}}
\newcommand{\ecor}{\end{cor}}
\newcommand{\fpro}{\end{proof}}
\newcommand{\eas}{\end{as}}
\newcommand{\ealg}{\end{alg}}

\theoremstyle{plain}

\newtheorem{theos}{Theorem}
\newtheorem{props}{Proposition}
\newtheorem{lems}{Lemma}
\newtheorem{cors}{Corollary}

\theoremstyle{definition}
\newtheorem{exas}{Example}
\newtheorem{algs}{Algorithm}
\newtheorem{asss}{Assumption}
\newtheorem{defns}{Definition}

\newcommand{\btheos}{\begin{theos}}
\newcommand{\etheos}{\end{theos}}
\newcommand{\bprops}{\begin{props}}
\newcommand{\eprops}{\end{props}}
\newcommand{\bdes}{\begin{defns}}
\newcommand{\edes}{\end{defns}}
\newcommand{\blems}{\begin{lems}}
\newcommand{\elems}{\end{lems}}
\newcommand{\bcors}{\begin{cors}}
\newcommand{\ecors}{\end{cors}}
\newcommand{\bexs}{\begin{exas}}
\newcommand{\eexs}{\end{exas}}
\newcommand{\balgs}{\begin{algs}}
\newcommand{\ealgs}{\end{algs}}
\newcommand{\bass}{\begin{asss}}
\newcommand{\eass}{\end{asss}}

%% file: variables.tex
\def \TaperCoef {v_{ij}}
\def \PrecCenterBlock {\Omega_l^{(m)}}
\def \PrecPaddedBlock {\Omega_{l-m}^{(3m)}}

%% file: intro.tex
Imposing structure is crucial to performing statistical estimation in
the high-dimensional regime where the number of observations can be
much smaller than the number of parameters.  In estimating graphical
models, a long line of work has focused on understanding how to impose
sparsity on the underlying graph structure.

Sparse edge recovery is generally not easy for an arbitrary
distribution. However, for Gaussian graphical models, it is well-known that the
graphical structure is encoded in the inverse of the covariance matrix
\(\Sigma^{-1} = \Omega\),
commonly referred to as the precision matrix \citep{lauritzen,
Meinshausen06, cai_constrained_2011}.  Therefore,
accurate recovery of the precision matrix is paramount to
understanding the structure of the graphical model.  As a consequence,
a great deal of work has focused on sparse recovery of precision
matrices under the multivariate normal
assumption~\cite{FriedHasTib2007,cai_estimating_2012,cai2016estimating,Rot09,ren_asymptotic_2015}. Beyond
revealing the graph structure, the precision matrix also turns out to
be highly useful in a variety of applications, including portfolio
optimization, speech recognition, and genomics~\cite{lauritzen,YuaLin07,5947493}.

Although there has been a rich literature exploring the sparse precision
matrix setting for Gaussian graphical models, less work has emphasized
understanding the estimation of precision matrices under additional
structural assumptions, with some exceptions for block structured
sparsity~\cite{hoslee16} or bandability~\cite{bicgel11}.
One would hope that extra structure should
allow us to obtain more statistically efficient solutions. In this
work, we focus on the case of bandable precision matrices, which capture
a sense of locality between variables. Bandable matrices arise in a
number of time-series contexts and have applications in climatology,
spectroscopy, fMRI analysis, and astronomy~\cite{fri94,vis95,pad16}.
For example, in the time-series setting, we may assume that edges
between variables \(X_i, X_j\)
are more likely when \(i\)
is temporally close to \(j\),
as is the case in an auto-regressive process.  The precision and
covariance matrices corresponding to distributions with this property
are referred to as bandable, or tapering. We will discuss the details
of this model in the sequel.

\paragraph{Past work:}
Previous work has explored the estimation of both bandable covariance and
precision matrices~\cite{cai_optimal_2010,pad16}.  Closely
related work includes the estimation of sparse precision and covariance
matrices~\cite{cai_constrained_2011,Rot09,cai_estimating_2012}. Asymptotically-normal entrywise precision
estimates as well as minimax rates for operator norm recovery of
sparse precision matrices have also been
established~\cite{ren_asymptotic_2015}. A line of work developed 
concurrently to our own establishes a matching minimax lower bound~\cite{leelee17}.

When considering an estimation technique, a powerful criterion for evaluating whether
the technique performs optimally in terms of convergence rate is
minimaxity.  Past work has established minimax rates of convergence
for sparse covariance matrices, bandable covariance matrices, and
sparse precision
matrices~\cite{CaiZhou10,cai_optimal_2010,cai_estimating_2012,Rot09}.

The technique for estimating bandable covariance matrices proposed
in~\cite{cai_optimal_2010} is shown to achieve the optimal rate of
convergence. However, no such theoretical guarantees have been shown for the bandable precision
estimator proposed in recent work for estimating sparse and smooth
precision matrices that arise from cosmological data~\cite{pad16}.

Of note is the fact that the minimax rate of convergence for
estimating sparse covariance matrices matches the minimax rate of
convergence of estimating sparse precision matrices.  In this paper,
we introduce an adaptive estimator and show that it achieves the
optimal rate of convergence when estimating bandable precision
matrices from the banded parameter space~\eqref{eq:parameter-space}.  We
find, satisfyingly, that analogous to the sparse case, in which the
minimax rate of convergence enjoys the same rate for both precision
and covariance matrices, the minimax rate of convergence for
estimating bandable precision matrices matches the minimax rate of
convergence for estimating bandable covariance matrices that has been
established in the literature~\cite{cai_optimal_2010}.

\paragraph{Our contributions:}
Our goal is to estimate a banded precision matrix based on $n$
i.i.d. observations. We consider a parameter space of precision matrices \(\Omega\) with a
power law decay structure nearly identical to the bandable covariance matrices
considered for covariance matrix estimation~\cite{cai_optimal_2010}.
We present a simple-to-implement algorithm for estimating the precision
matrix.  Furthermore, we show that the algorithm is minimax optimal with
respect to the spectral norm.
The upper and
lower bounds given in Section \ref{sect:optimality} together imply the
following optimal rate of convergence for estimating bandable precision
matrices under the spectral norm. Informally, our results show the
following bound for recovering a banded precision matrix with bandwidth
$k$.
\begin{theorem}[Informal]
    The minimax risk for estimating the precision matrix \(\Omega\) over
    the class \(\Pp_\alpha\) given in (\ref{eq:parameter-space}) satisfies:
    \begin{equation}
        \inf_{\hat\Omega} \sup_{\Pp_\alpha}
        \EE\norm{\hat\Omega - \Omega}^2
        \approx
        \frac{k+\log p}{n}
    \end{equation}
    where this bound is achieved by
    the tapering estimator
    \(\hat\Omega_k\) as defined in Equation~\eqref{eq:tapering-estimator}.
\end{theorem}
An important point to note, which is shown more precisely in the
sequel, is that the rate of convergence as compared to sparse
precision matrix recovery is improved by a factor of $\min(k \log(p),k^2)$.

We establish a minimax upper bound by detailing an algorithm for obtaining
an estimator given observations \(\xx_1, \dotsc, \xx_n\) and a pre-specified
bandwidth \(k\), and studying the resultant estimator's risk properties under
the spectral norm.  We show that an estimator using our algorithm with the
optimal choice of bandwidth attains the minimax rate of convergence with high
probability.

To establish the optimality of our estimation routine, we derive a minimax
lower bound to show that the rate of convergence cannot be improved beyond
that of our estimator.  The lower bound is established by constructing
subparameter spaces of (\ref{eq:parameter-space}) and applying testing
arguments through Le Cam's method and Assouad's Lemma~\cite{Yu97,cai_optimal_2010}.

To supplement our analysis, we conduct numerical experiments to
explore the performance of our estimator in the finite sample setting.  The
numerical experiments confirm that even in the finite sample case, our
proposed estimator exhibits the minimax rate of convergence.

The remainder of the paper is organized as follows.  In Section
\ref{sect:methodology}, we detail the exact model setting and
introduce a blockwise inversion technique for precision matrix
estimation.  In Section \ref{sect:optimality}, theorems establishing
the minimaxity of our estimator under the spectral norm are presented.
An upper bound on the estimator's risk is given in high probability
with the help of a result from set packing.  The minimax lower bound
is derived by way of a testing argument.  Both bounds are accompanied
by their proofs.  Finally, in Section \ref{sect:simulations}, our
estimator is subjected to numerical experiments.  Owing to space
constraints, proofs for auxiliary lemmas may be found in Appendix
\ref{aux-lemma-proofs}.

\paragraph{Notation:}
We will now collect notation that will be used
throughout the remaining sections. Vectors will be denoted as
lower-case $\xx$ while matrices are upper-case $A$. The spectral or operator norm
of a matrix is defined to be $\norm{A} = \sup_{\xx \neq 0,\yy \neq 0}
\langle A \xx , \yy \rangle$ while the matrix $\ell_1$ norm of a
symmetric matrix $A \in \RR^{m \times m}$ is defined to be $\norm{A}_1
= \max_{j} \sum_{i=1}^m |A_{ij}|$.

%% file: methodology.tex
In this section we present details of our model and the estimation procedure.  If one considers observations of the form
\(\xx_1, \dotsc, \xx_n \in \RR^p\) drawn from a distribution with precision
matrix \(\Omega_{p\times p}\) and zero mean, the goal then is to estimate the
unknown matrix \(\Omega_{p\times p}\) based on the observations
\(\{\xx_i\}_{i=1}^n\). Given a random sample of \(p\)-variate observations
\(\xx_1, \dotsc, \xx_n\) drawn from a multivariate distribution with
population covariance \(\Sigma = \Sigma_{p \times p}\), our procedure is
based on a tapering estimator derived from blockwise estimates for
estimating the precision matrix \(\Omega_{p \times p} = \Sigma^{-1}\).

The maximum likelihood estimator of \(\Sigma\) is
\begin{equation}\label{eq:cov-mle}
  \hat\Sigma
  = (\hat\sigma_{ij})_{1 \leq i, j \leq p}
  = \frac{1}{n} \sum_{l=1}^n (\xx_l - \bar \xx)(\xx_l - \bar \xx)^\top
\end{equation}
where $\bar \xx$ is the empirical mean of the vectors $\xx_i$. We will construct estimators of the precision matrix \(\Omega = \Sigma^{-1}\)
by inverting blocks of $\hat\Sigma$ along the diagonal, and averaging over the
resultant subblocks.

Throughout this paper we adhere to the convention that \(\omega_{ij}\) refers
to the \(ij^\text{th}\) element in a matrix \(\Omega\).  Consider the parameter
space \(\Ff_\alpha\), with associated probability measure \(\Pp_\alpha\), given
by: \begin{equation}\label{eq:parameter-space}
    \Ff_\alpha
    = \Ff_\alpha(M_0, M) =
    \Bigg\{
        \Omega: \max_j \sum_i
        \left\{
            |\omega_{ij}|: |i - j| \geq k
        \right\}
        \leq Mk^{-\alpha} \text{ for all } k,
        \lambda_i(\Omega) \in [M_0^{-1}, M_0]
    \Bigg\}
\end{equation}
where \(\lambda_i(\Omega)\) denotes the \(i\)th eigenvalue of \(\Omega\), with
\(\lambda_i \geq \lambda_j\) for all \(i \leq j\).  We also constrain
\(\alpha > 0, M > 0, M_0 > 0\).  Observe that this parameter space is nearly
identical to that given in Equation (3) of \citep{cai_optimal_2010}.  We take
on an additional assumption on the minimum eigenvalue of
\(\Omega \in \Ff_\alpha\), which is used in the technical arguments where the
risk of estimating \(\Omega\) under the spectral norm is bounded in terms of
the error of estimating \(\Sigma = \Omega^{-1}\).

Observe that the parameter space intuitively dictates that the magnitude of
the entries of \(\Omega\) decays in power law as we move away from the
diagonal.  As with the parameter space for bandable covariance matrices given
in \citep{cai_optimal_2010}, we may understand \(\alpha\) in
(\ref{eq:parameter-space}) as a rate of decay for the precision entries
\(\omega_{ij}\) as they move away from the diagonal; it can also be understood
in terms of the smoothness parameter in nonparametric estimation
\citep{Tsybakov:2008:INE:1522486}.  As will be discussed in Section
\ref{sect:optimality}, the optimal choice of \(k\) depends on both \(n\)
and the decay rate \(\alpha\).

\subsection{Estimation procedure}\label{subsect:estimation-procedure}
We now detail the algorithm for obtaining minimax estimates for bandable
\(\Omega\), which is also given as pseudo-code\footnote{
    In the pseudo-code, we adhere to the NumPy convention (1) that arrays are
    zero-indexed, (2) that slicing an array \texttt{arr} with the operation
    \texttt{arr[a:b]} includes the element indexed at \texttt{a} and excludes
    the element indexed at \texttt{b}, and (3) that if \texttt{b} is greater
    than the length of the array, only elements up to the terminal element are
    included, with no errors.
}
in Algorithm \ref{alg:blockwise}. 

The algorithm is inspired by the tapering procedure introduced by
Cai, Zhang, and Zhou~\cite{cai_optimal_2010} in the case of covariance
matrices, with modifications in order to estimate the
precision matrix. Estimating the precision matrix introduces new difficulties as we do not have
direct access to the estimates of elements of the precision
matrix. For a given integer \(k, 1 \leq k \leq p\),
we construct a tapering estimator as follows.  First, we calculate the
maximum likelihood estimator for the covariance, as given in Equation
(\ref{eq:cov-mle}).  Then, for all integers \(1 - m \leq l \leq p\)
and \(m \geq 1\),
we define the matrices with square blocks of size at most \(3m\)
along the diagonal:
\begin{equation}\label{eq:cov-subblock}
     \hat\Sigma_{l - m}^{(3m)} =
    (\hat\sigma_{ij}
    \one\{l  - m\leq i < l + 2m, l - m\leq j < l + 2m\})_{p \times p}
\end{equation}
For each \(\hat\Sigma_{l - m}^{(3m)}\), we replace the
nonzero block with its inverse to obtain \(\breve\Omega_{l-m}^{(3m)}\).  For a
given \(l\), we refer to the individual entries of this intermediate matrix as
follows:
\begin{equation}\label{eq:prec-big-subblock}
    \breve\Omega_{l-m}^{(3m)} = 
    (\breve\omega^l_{ij}
    \one\{l - m \leq i < l + 2m, l - m\leq j < l + 2m\})_{p \times p}
\end{equation}
For each \(l\), we then keep only the central \(m \times m\) subblock of
\(\breve\Omega_{l-m}^{(3m)}\) to obtain the blockwise estimate
\(\hat\Omega_l^{(m)}\): 
\begin{equation}\label{eq:prec-subblock}
    \hat\Omega_l^{(m)} = 
    (\breve\omega^l_{ij}
    \one\{l \leq i < l + m, l \leq j < l + m\})_{p \times p}
\end{equation}
Note that this notation allows for \(l < 0\) and \(l + m > p\); in each case,
this out-of-bounds indexing allows us to cleanly handle corner cases where the
subblocks are smaller than \(m\times m\).

For a given bandwidth \(k\) (assume \(k\) is divisible by 2), we calculate
these blockwise estimates for both \(m = k\) and \(m =
\frac{k}{2}\). Finally, we construct our estimator by averaging over the block matrices:
\begin{equation}\label{eq:tapering-estimator}
    \hat\Omega_k = \frac{2}{k} \cdot \left(
        \sum_{l = 1 - k}^p \hat\Omega_l^{(k)}
        - \sum_{l = 1 - \nicefrac{k}{2}}^p \hat\Omega_l^{(\nicefrac{k}{2})}
    \right)
\end{equation}
We note that within \(\frac{k}{2}\) entries of the diagonal, each entry
is effectively the sum of \(\frac{k}{2}\) estimates, and as we move from
\(\frac{k}{2}\) to \(k\) from the diagonal, each entry is progressively the sum
of one fewer entry.

Therefore, within \(\frac{k}{2}\) of the diagonal, the entries are not tapered;
and from \(\frac{k}{2}\) to \(k\) of the diagonal, the entries are linearly
tapered to zero.  The analysis of this estimator makes careful use of this
tapering schedule and the fact that our estimator is constructed through the
average of block matrices of size at most \(k \times k\).

\begin{algorithm}
    \begin{algorithmic}
        \Function{FitBlockwise}{\(\hat\Sigma\), k}
            \State \(\hat\Omega \gets \mathbf{0}_{p\times p}\)
            \For{\(l \in [1-k, p)\)}
                \State \(\hat\Omega \gets \hat\Omega
                    + \Call{BlockInverse}{\hat\Sigma, k, l}\)
            \EndFor
            \For{\(l \in [1-\lfloor \nicefrac{k}{2}\rfloor, p)\)}
                \State \(\hat\Omega \gets \hat\Omega
                    - \Call{BlockInverse}{\hat\Sigma,
                        \lfloor\nicefrac{k}{2}\rfloor, l}\)
            \EndFor
            \State \Return \(\hat\Omega\)
        \EndFunction
        \State
        \Function{BlockInverse}{\(\hat\Sigma\), m, l}
            \State
            \Comment{Obtain \(3m \times 3m\) block inverse.}
            \State \(s \gets \max\{l - m, 0\}\)
            \State \(f \gets \min\{p, l + 2m\}\)
            \State \(M \gets \left(\hat\Sigma\texttt{[s:f, s:f]}\right)^{-1}\)
            \State
            \Comment{Preserve central \(m \times m\) block of inverse.}
            \State \(s \gets m + \min\{l - m, 0\}\)
            \State \(N \gets M\texttt{[s:s+m,
                    s:s+m]}\)
            \State
            \Comment{Restore block inverse to appropriate indices.}
            \State \(s \gets \max\{l, 0\}\)
            \State \(f \gets \min\{l + m, p\}\)
            \State \(P\texttt{[s:f, s:f]} = N\)
            \State\Return \(P\)
        \EndFunction
    \end{algorithmic}
    \caption{Blockwise Inversion Technique}\label{alg:blockwise}
\end{algorithm}

\subsection{Implementation details}\label{subsect:implem-details}
The naive algorithm performs \(O(p + k)\)
inversions of square matrices with size at most \(3k\).
This method can be sped up considerably through an application of the
Woodbury matrix identity and the Schur complement
relation~\cite{MR0038136,Boyd02}. Doing so reduces the computational
complexity of the algorithm from \(O(pk^3)\)
to \(O(pk^2)\).
We discuss the details of modified algorithm and its computational
complexity below.

Suppose we have \(\breve\Omega_{l-m}^{(3m)}\)
and are interested in obtaining \(\breve\Omega_{l-m+1}^{(3m)}\).
We observe that the nonzero block of \(\breve\Omega_{l-m+1}^{(3m)}\)
corresponds to the inverse of the nonzero block of
\(\hat\Sigma_{l - m +1}^{(3m)}\),
which only differs by one row and one column from
\(\hat\Sigma_{l - m}^{(3m)}\),
the matrix for which the inverse of the nonzero block corresponds to
\(\breve\Omega_{l-m}^{(3m)}\),
which we have already computed. We may understand the movement from
\(\hat\Sigma_{l - m}^{(3m)}, \breve \Omega_{l-m}^{(3m)}\)
to \(\hat\Sigma_{l - m + 1}^{(3m)}\)
(to which we already have direct access) and
\(\breve\Omega_{l-m+1}^{(3m)}\)
as two rank-1 updates.  Let us view the nonzero blocks of
\(\hat\Sigma_{l - m}^{(3m)}, \breve\Omega_{l-m}^{(3m)}\) as the block matrices:
\begin{align*}
    \mathrm{NonZero}(\hat\Sigma_{l - m}^{3m}) &= \bmat{
        A \in \RR^{1\times 1}    &   B \in \RR^{1 \times (3m - 1)}   \\ 
        B^\top \in \RR^{(3m - 1) \times 1} & C \in \RR^{(3m - 1) \times (3m -
            1)} 
    }   \\
    \mathrm{NonZero}(\breve\Omega_{l-m}^{(3m)}) &= \bmat{
        \tilde A \in \RR^{1\times 1}    &   \tilde B \in \RR^{1 \times (3m - 1)}
        \\ 
        \tilde B^\top \in \RR^{(3m - 1) \times 1} & \tilde C \in \RR^{(3m - 1)
            \times (3m - 1)} 
    }
\end{align*}
The Schur complement relation tells us that given \(\hat\Sigma_{l - m}^{3m},
    \breve\Omega_{l-m}^{(3m)}\), we may trivially compute \(C^{-1}\) as follows:
\begin{equation}\label{eq:schur-update}
    C^{-1} = \left(\tilde C^{-1} + B^\top A^{-1} B\right)^{-1} 
    =  \tilde C - \frac{\tilde C B^\top B\tilde C}{A + B \tilde CB^\top}
\end{equation}
by the Woodbury matrix identity, which gives an efficient algorithm for
computing the inverse of a matrix subject to a low-rank (in this case, rank-1)
perturbation.  This allows us to move from the inverse of a matrix in
\(\RR^{3m \times 3m}\) to the inverse of a matrix in
\(\RR^{(3m-1)\times(3m-1)}\) where a row and column have been removed.  A
nearly identical argument allows us to move from the \(\RR^{(3m-1)\times(3m-1)}\)
matrix to an \(\RR^{3m \times 3m}\) matrix where a row and column have been
appended, which gives us the desired block of \(\breve\Omega_{l-m+1}^{(3m)}\).

With this modification to the algorithm, we need only compute the inverse of a
square matrix of width \(2m\) at the beginning of the routine; thereafter,
every subsequent block inverse may be computed through simple rank one
matrix updates.

\subsection{Complexity details}
We now detail the factor of \(k\) improvement in computational complexity provided through
the application of the Woodbury matrix identity and the Schur complement
relation introduced in Section \ref{subsect:implem-details}.  Recall that the
naive implementation of Algorithm \ref{alg:blockwise} involves \(O(p+k)\)
inversions of square matrices of size at most \(3k\), each of which cost
\(O(k^3)\).  Therefore, the overall complexity of the naive algorithm is
\(O(pk^3)\), as \(k < p\).

Now, consider the Woodbury-Schur-improved algorithm.  The initial single inversion of a
\(2k \times 2k\) matrix costs \(O(k^3)\).  Thereafter, we perform \(O(p + k)\)
updates of the form given in Equation (\ref{eq:schur-update}).  These
updates simply require vector matrix operations. Therefore, the update
complexity on each iteration is \(O(k^2)\).  It follows that the overall
complexity of the amended algorithm is \(O(pk^2)\).

%% file: optimality.tex
Here we present the results that establish the rate optimality of the
above estimator under the spectral norm. For symmetric matrices \(A\), the spectral norm, which
corresponds to the largest singular value of \(A\), coincides with the
\(\ell_2\)-operator norm.  We establish optimality by first deriving an
upper bound in high probability using the blockwise inversion estimator defined
in Section~\ref{subsect:estimation-procedure}.  We then give a matching
lower bound in expectation by carefully constructing two sets of multivariate
normal distributions and then applying Assouad's Lemma and Le Cam's
method.

\subsection{Upper bound under the spectral norm}
\label{subsect:upper-bound}
In this section we derive a risk upper bound for the tapering estimator defined
in (\ref{eq:tapering-estimator}) under the operator norm.  We assume the
distribution of the \(\xx_i\)'s is subgaussian; that is, there exists
\(\rho > 0\) such that:
\begin{equation}\label{eq:subgaussian}
    \PP\left\{
        |\vv^\top(\xx_i - \EE\xx_i)| > t
    \right\} \leq e^{-\frac{t^2\rho}{2}}
\end{equation}
for all \(t > 0\) and \(\norm{\vv}_2 = 1\).  Let
\(\Pp_\alpha = \Pp_\alpha(M_0, M, \rho)\) denote the set of distributions of
\(\xx_i\) that satisfy (\ref{eq:parameter-space}) and (\ref{eq:subgaussian}).

\begin{theorem}\label{thm:upper-bound}
    The tapering estimator \(\hat\Omega_k\), defined in
    \eqref{eq:tapering-estimator}, of the precision matrix
    \(\Omega_{p\times p}\) with \(p > n^\frac{1}{2\alpha + 1}\) satisfies:
    \begin{equation}\label{eq:upper-bound-elementary}
        \sup_{\Pp_\alpha}\PP\left\{
            \norm{\hat\Omega_k - \Omega}^2
            \geq
            C\frac{k + \log p}{n} + Ck^{-2\alpha}
        \right\}
        = O\left(p^{-15}\right)
    \end{equation}
    with \(k = o(n)\), \(\log p = o(n)\), and a universal constant \(C > 0\). 

    In particular, the estimator \(\hat\Omega = \hat\Omega_k\) with
    \(k = n^\frac{1}{2\alpha+1}\) satisfies:
    \begin{equation}\label{eq:upper-bound-optimal}
        \sup_{\Pp_\alpha}\PP\left\{
            \norm{\hat\Omega_k - \Omega}^2
            \geq Cn^{-\frac{2\alpha}{2\alpha+1}} + C\frac{\log p}{n}
        \right\}
        = O\left(p^{-15}\right)
    \end{equation}
\end{theorem}

Given the result in Equation (\ref{eq:upper-bound-elementary}), it is easy
to show that setting \(k = n^\frac{1}{2\alpha+1}\) yields the optimal rate
by balancing the size of the inside-taper and outside-taper terms, which gives
Equation (\ref{eq:upper-bound-optimal}).

The proof of this theorem, which is given next, relies
on the fact that when we invert a \(3k \times 3k\) block, the difference
between the central \(k \times k\) block and the corresponding \(k \times k\) block
which would have been obtained by inverting the full matrix has a negligible
contribution to the risk.  As a result, we are able to take concentration
bounds on the operator norm of subgaussian matrices, customarily used for
bounding the norm of the difference of covariance matrices, and apply them
instead to differences of precision matrices to obtain our result.

The key insight is that we can relate the spectral norm of a
\(k \times k\)
subblock produced by our estimator to the spectral norm of the
corresponding \(k \times k\)
subblock of the covariance matrix, which allows us to apply
concentration bounds from classical random matrix theory.  Moreover,
it turns out that if we apply the tapering schedule induced by the
construction of our estimator to the population parameter
\(\Omega \in \Ff_\alpha\),
we may express the tapered population \(\Omega\)
as a sum of block matrices in exactly the same way that our estimator
is expressed as a sum of block matrices.

In particular, the tapering schedule is presented next. Suppose a population
precision matrix \(\Omega \in \Ff_\alpha\).  Then, we denote the tapered
version of \(\Omega\) by \(\Omega_A\), and construct:
\begin{align*}
    \Omega_A    &=  \left(\omega_{ij}\cdot \TaperCoef\right)_{p\times p} \\ 
    \Omega_B    &=  \left(\omega_{ij}\cdot(1 - \TaperCoef\right))_{p\times p} \\
\end{align*}
where the tapering coefficients are given by:
\begin{align*}
    \TaperCoef = \begin{cases}
        1   \ffor   |i - j| < \frac{k}{2}   \\
        \frac{|i-j|}{\nicefrac{k}{2}}   \ffor   \frac{k}{2} \leq |i - j| < k   \\
        0   \ffor   |i - j| \geq k   \\
    \end{cases}
\end{align*}
We then handle the risk of estimating the inside-taper \(\Omega_A\) and the
risk of estimating the outside-taper \(\Omega_B\) separately.

Because our estimator and the population parameter are both averages over \(k
    \times k\) block matrices along the diagonal, we may then take a union
bound over the high probability bounds on the spectral norm deviation for
the \(k \times k\) subblocks to obtain a high probability bound on the risk of
our estimator.

\subsubsection{Proof of Theorem \ref{thm:upper-bound}}
\label{proof:thm-upper-bound}

The main step in proving the upper bound on the estimation rate is
bounding the error for a tapered version of the truth and its
complement separately.  Let us denote a tapering coefficient:
\begin{equation}\label{eq:tapering-coef}
    \TaperCoef = \begin{cases}
        1   \ffor   |i - j| < \frac{k}{2}   \\
        \frac{|i-j|}{\nicefrac{k}{2}}   \ffor   \frac{k}{2} \leq |i - j| < k   \\
        0   \ffor   |i - j| \geq k   \\
    \end{cases}
\end{equation}
Let us denote:
\begin{align*}
    \Omega_A    &=  \left(\omega_{ij}\cdot \TaperCoef\right)_{p\times p} \\ 
    \Omega_B    &=  \left(\omega_{ij}\cdot(1 - \TaperCoef\right))_{p\times p} \\
\end{align*}
We similarly decompose:
\begin{align*}
    \hat\Omega_A    &=  \left(\hat\omega_{ij}\cdot\one\{|i - j| <
    k\}\right)_{p\times p} \\ 
    \hat\Omega_B    &=  \left(\hat\omega_{ij}\cdot\one\{|i - j| \geq
    k\}\right)_{p\times p} \\ 
\end{align*}

We will first show that the error against the tapered truth satisfies:
\begin{equation}\label{eq:error-inside-band}
    \PP\left\{
        \norm{\hat\Omega_A - \Omega_A}^2
        \geq
        C\left(\frac{k + \log p}{n}\right)
        + Ck^{-4a}
    \right\}
    = O\left(p^{-15}\right)
\end{equation}
and that the error outside the taper satisfies the deterministic bound:
\begin{equation}\label{eq:error-outside-band}
    \norm{\hat\Omega_B - \Omega_B}^2 \leq Ck^{-2\alpha}
\end{equation}
It then follows that:
\begin{align*}
    \PP&\left\{
        \norm{\hat\Omega - \Omega}^2
        \geq 
        C\frac{k + \log p}{n} + Ck^{-2\alpha}
    \right\}    \\
    &\leq  
    \PP\left\{
        2\norm{\hat\Omega_A - \Omega_A}^2
        + 2\norm{\hat\Omega_B - \Omega_B}^2
        \geq 
        C\frac{k + \log p}{n} + Ck^{-4\alpha}  + Ck^{-2\alpha}
    \right\} \\
    &\leq 
    \PP\left\{
        \norm{\hat\Omega_A - \Omega_A}^2
        \geq
        C\frac{k + \log p}{n} + Ck^{-4\alpha} 
    \right\} +
    \PP\left\{
        \norm{\hat\Omega_B - \Omega_B}^2
        \geq 
        Ck^{-2\alpha}
    \right\} \\
    &= O\left(p^{-15}\right)
\end{align*}
This proves (\ref{eq:upper-bound-elementary}), from which follows
(\ref{eq:upper-bound-optimal}).  Therefore, the estimator $\hat\Omega$
with $k = n^\frac{1}{2\alpha+1}$ satisfies:
\begin{align*}
    \PP\left\{
        \norm{\hat\Omega - \Omega}^2
        \leq
        2C\left( n^{-\frac{2\alpha}{2\alpha+1}} + \frac{\log p}{n}\right)
    \right\}
    &= O\left(p^{-15}\right)
\end{align*}
This proves Theorem \ref{thm:upper-bound}.

We first establish (\ref{eq:error-outside-band}), which is relatively simple.
Observe that by definition, $\hat\Omega_B$ is the zero matrix, as
$\hat\Omega$ already sets all entries outside the band to zero.  Therefore:
\begin{align*}
    \norm{\hat\Omega_B - \Omega_B}^2
    &=  \norm{\Omega_B}^2   \\
    &\leq \norm{\Omega_B}_1^2   \\
    &=     \left[
        \max_j \sum_{i} |\omega_{ij}\cdot(1 - \TaperCoef)|
        \right]^2   \\
    &\leq  \left[
        \max_j \sum_{|i - j| > \frac{k}{2}} |\omega_{ij}|
        \right]^2   \\
    &\leq \left[ M2^\alpha k^{-\alpha}\right]^2 \\
    &= Ck^{-2\alpha}
\end{align*}

We now show (\ref{eq:error-inside-band}). Let
\(
    \Omega_l^{(m)} = (\omega_{ij}
    \one\{l \leq i < l + m, l \leq j < l + m\})_{p \times p}
\). 
\begin{lemma}\label{lemma:population-sum-of-blocks}
    We may express the tapered population parameter as:
    \begin{equation}\label{eq:tapered-population-param}
        \Omega_A = \frac{2}{k} \cdot \left(
            \sum_{l = 1 - k}^p \Omega_l^{(k)}
            - \sum_{l = 1 - \nicefrac{k}{2}}^p \Omega_l^{(\nicefrac{k}{2})}
        \right)
    \end{equation}
\end{lemma}
Then define:
\begin{equation}
    N^{(m)} = \max_{1-m \leq l \leq p} \norm{
        \hat\Omega_l^{(m)} - \Omega_l^{(m)}
    }
\end{equation}

\begin{lemma}\label{lemma:bound-by-max}
    Let \(\hat\Omega = \hat\Omega_m\) be defined as in
    \eqref{eq:tapering-estimator}.  Then
    \begin{align*}
        \norm{\hat\Omega - \Omega_A} \leq C\cdot N^{(m)}
    \end{align*}
\end{lemma}

We then show that our estimation technique approximates each block of the
true precision matrix up to a lower order correction.

\begin{lemma}\label{lemma:schur-correction}
    The \(m \times m\) block of $\Omega$ starting at the \(l\)th diagonal
    entry may be expressed as an approximation \(\tilde\Omega_l^{(m)}\) from
    inverting blocks of the covariance matrix \(\Sigma\) plus a correction
    term \(W_l^{(m)}\).
    \begin{equation}
        \Omega_l^{(m)} = \tilde \Omega_l^{(m)} + W_l^{(m)}
    \end{equation}
    In particular, \(W_l^{(m)}\) takes the form:
    \begin{align*}
        W_l^{(m)} = \Omega_{B_2}\Omega_C^{-1}\Omega_{B_2}^\top
    \end{align*}
    with:
    \begin{align*}
        \Omega_{B_2}
        &= \bmat{\Omega_\alpha  &   \Omega_\beta}
        \\
        \Omega_C
        &=  \bmat{\Omega_\gamma & \Omega_\delta^\top
            \\ \Omega_{\delta} & \Omega_\epsilon}
    \end{align*}
    where we define the block matrices:
    \begin{align*}
        \Omega_\alpha
        &= \Omega_{l \leq i < l + m, l + 2m \leq j \leq p}
        \\
        \Omega_\beta
        &= \Omega_{l \leq i < l + m, 1 \leq j < l - m}
        \\
        \Omega_\gamma
        &= \Omega_{l + 2m \leq i, j \leq p}
        \\
        \Omega_\delta
        &= \Omega_{l - m \leq i < l + 2m, 1 \leq j < l - m}
        \\
        \Omega_\epsilon
        &= \Omega_{1 \leq i, j < l - m}
        \\
    \end{align*}
    and \(\tilde\Omega_l^{(m)}\) is given by the central \(m \times m\) block
    of \(\Sigma_{l - m}^{(3m)^{-1}}\).
\end{lemma}

\begin{lemma}\label{lemma:correction-rate}
    The correction factor \(W\) in Lemma \ref{lemma:schur-correction} is
    bounded in spectral norm:
    \begin{equation}
        \norm{W_l^{(m)}} \leq Cm^{-2\alpha}
    \end{equation}
\end{lemma}

We may then control the operator norm of each $m \times m$ random matrix with
$m = k$ as follows.  First, we bound \(N^{(m)}\) from above by two terms:
\begin{align*}
    N^{(m)}
    &=  \max_{1 - m \leq l \leq p} \norm{
        \hat\Omega_l^{(m)} - \Omega_l^{(m)}
    }   \\
    &\leq \underbrace{\max_{1 - m \leq l \leq p} \norm{
        \hat\Omega_l^{(m)} - \tilde\Omega_l^{(m)}
    }}_{N_1^{(m)}} +
    \underbrace{\max_{1 - m \leq l \leq p}
    \norm{
        \tilde\Omega_l^{(m)} - \Omega_l^{(m)}
    }}_{N_2^{(m)}}
\end{align*}
Note that \(N_2^{(m)} = \max_l \norm{W_l^{(m)}}\).  Therefore, we already have
a deterministic bound on \(N_2^{(m)}\) from Lemma \ref{lemma:correction-rate}.

Using standard results from random matrix theory we may bound
$N_1^{(m)}$  with high probability in the following lemma. We defer
the proof to the Appendix.
\begin{lemma}\label{lemma:max-conc-bound}
    There exists a constant \(\rho_1 > 0\) such that:
    \begin{equation}\label{eq:max-conc-bound}
        \PP\left\{
            N_1^{(m)} > x
        \right\}
        \leq
        2p\cdot25^{3m}\exp\left\{
        -nx^2\rho_1
        \right\}
    \end{equation}
    for all \(0 < x < \rho_1\) and \(1 - m \leq l \leq p\).
\end{lemma}

We now prove the upper bound in high probability on the within-band error in
Equation (\ref{eq:error-inside-band}).  First, by setting \(x =
  4\sqrt{\frac{\log p + m}{n\rho_1}}\), and recalling Lemma
\ref{lemma:max-conc-bound}, we have:
\begin{align*}
    \PP\left\{
        \left(N_1^{(m)}\right)^2
        \geq
        16\frac{\log p + m}{n\rho_1}
    \right\}
    &\leq
    2p\cdot25^{3m}\exp\left\{
        -16\log p - 16 m
    \right\}    \\
\end{align*}
This immediately implies that:
\begin{align*}
    \PP\left\{
        \left(N_1^{(m)}\right)^2
        \geq
        C\frac{\log p + m}{n}
    \right\}
    &=  O\left(p^{-15}\right)
\end{align*}
Finally, we apply Lemmas \ref{lemma:bound-by-max} and \ref{lemma:correction-rate}:
\begin{align*}
    \PP&\left\{
        \norm{\hat\Omega_A - \Omega_A}^2
        \geq
        C\left(\frac{k + \log p}{n}\right)
        + Ck^{-4a}
    \right\}    \\
    &\leq
    \PP\left\{
        \left(N^{(m)}\right)^2
        \geq
        C\left(\frac{k + \log p}{n}\right)
        + Ck^{-4a}
    \right\}    \\
    &\leq
    \PP\left\{
        2\left(N_1^{(m)}\right)^2 + 
        2\left(N_2^{(m)}\right)^2 
        \geq
        C\left(\frac{k + \log p}{n}\right)
        + Ck^{-4a}
    \right\}    \\
    &\leq
    \PP\left\{
        2\left(N_1^{(m)}\right)^2
        \geq
        C\left(\frac{k + \log p}{n}\right)
    \right\}
    +
    \PP\left\{
        2\left(N_2^{(m)}\right)^2 
        \geq
        Ck^{-4a}
    \right\}    \\
    &= \PP\left\{
        2\left(N_1^{(m)}\right)^2
        \geq
        C\left(\frac{k + \log p}{n}\right)
    \right\}    \\
    &= O\left(p^{-15}\right)
\end{align*}
which shows (\ref{eq:error-inside-band}).
\qed

\subsection{Lower bound under the spectral norm}\label{subsect:lower-bound}
In Section \ref{subsect:upper-bound}, we established Theorem
\ref{thm:upper-bound}, which states that our estimator achieves the rate of
convergence \(n^{-\frac{2\alpha}{2\alpha+1}}\) under the spectral norm by using
the optimal choice of \(k = n^\frac{1}{2\alpha+1}\).  Next we demonstrate a
matching lower bound, which implies that the upper bound established in
Equation (\ref{eq:upper-bound-optimal}) is tight up to constant factors.

Specifically, for the estimation of precision matrices in the parameter space
given by Equation (\ref{eq:parameter-space}), the following minimax lower
bound holds.

\begin{theorem}\label{thm:lower-bound}
    The minimax risk for estimating the precision matrix \(\Omega\) over
    \(\Pp_\alpha\) under the operator norm satisfies:
    \begin{equation}\label{eq:lower-bound}
        \inf_{\hat\Omega} \sup_{\Pp_\alpha} \EE\norm{\hat\Omega - \Omega}^2
        \geq cn^{-\frac{2\alpha}{2\alpha+1}} + c \frac{\log p}{n}
    \end{equation}
\end{theorem}

As in many information theoretic lower bounds, our first step is to
construct a set of multivariate normal
distributions; then we compute the total variation affinity between pairs of
probability measures in the set.

We will now select a subset of our parameter space that captures most
of the complexity of the full space. We then establish an information
theoretic limit on estimating parameters from this subspace, which
yields a valid minimax lower bound over the original set. Therefore, to
establish the lower bound given in Theorem \ref{thm:lower-bound}, we
construct two subparameter spaces, \(\Ff_{11}\)
and \(\Ff_{12}\),
and derive a lower bound on the estimation of precision matrices in
each set separately.  We then take the union
\(\Ff_1 = \Ff_{11} \cup \Ff_{12}\)
of the two subparameter spaces, and Equation (\ref{eq:lower-bound})
follows.

\paragraph{Subparameter space construction:}
We apply a similar technique as in the work for bounding the spectral norm
error for estimating covariance matrices~\cite{cai_optimal_2010}, with
adaptations for the precision matrix setting.

Given positive integers $k$ and $m$ such that $2k \leq p$ and $1 \leq m \leq
k$, we parameterize a set of matrices $B(m, k) = (b_{ij})_{p\times p}$ as:
\begin{align*}
    b_{ij} = \one\left\{ i = m \text{ and }m+1\leq j \leq 2k, \text{ or }
    j = m \text{ and } m + 1\leq i \leq 2k\right\}
\end{align*}

Let $k = n^\frac{1}{2\alpha + 1}$ and $a = k^{-\alpha - 1}$.  Then, we define
the following set of $2^k$ precision matrices, each parameterized by $\theta
\in \{0, 1\}^k$:
\begin{align*}
    \Ff_{11} = \left\{
    \Omega(\theta): \Omega(\theta) = \II_{p\times p}
    + \tau \alpha \sum_{m=1}^k \theta_m B(m, k)
    \right\}
\end{align*}
with $0 < \tau < 2^{-\alpha - 1}M$.  To this parameter space \(\Ff_{11}\), we
apply Assouad's Lemma to obtain a lower bound with rate
\(n^{-\frac{2\alpha}{2\alpha+1}}\).

Separately, we construct the subparameter space \(\Ff_{12}\) consisting of
diagonal matrices:
\[
    \Ff_{12} = \left\{
    \Omega_m = \omega_{ij} = \one\{i = j\}\left(
    1 + \one\{i = j = m\}\sqrt{\frac{\tau}{n}\log p_1}
    \right)^{-1}, 0 \leq m \leq p_1
    \right\}
\]
where $p_1 = \min\{p, \exp\{\frac{n}{2}\}\}$ and $0 < \tau < \min\{(M_0 - 1)^2,
(\rho - 1)^2, 1\}$.  To \(\Ff_{12}\), we apply Le Cam's method to obtain
a lower bound with rate \(\frac{\log p}{n}\).

\subsubsection{Proof of Theorem \ref{thm:lower-bound}}
\label{proof:thm-lower-bound}
Our proof strategy is as follows.  We will define two subparameter spaces
$\Ff_{11}, \Ff_{12} \subset \Ff_\alpha$, and prove a lower bound on the
estimation rate for each one.  More specifically, we will show that:
\begin{equation}\label{eq:F11-lower-bound}
    \inf_{\hat\Omega}\sup_{\Ff_{11}} \EE\norm{\hat\Omega - \Omega}^2
    \geq cn^{-\frac{2\alpha}{2\alpha+1}}
\end{equation}
and
\begin{equation}\label{eq:F12-lower-bound}
    \inf_{\hat\Omega}\sup_{\Ff_{12}} \EE\norm{\hat\Omega - \Omega}^2
    \geq c\frac{\log p}{n}
\end{equation}
for some constant $c > 0$.  Let $\Ff_1 = \Ff_{11} \cup \Ff_{12} \subset
\Ff_\alpha$.  Equations (\ref{eq:F11-lower-bound}) and
(\ref{eq:F12-lower-bound}) then together imply:
\begin{equation}
    \inf_{\hat\Omega}\sup_{\Ff_1} \EE\norm{\hat\Omega - \Omega}^2
    \geq \frac{c}{2}\left( n^{-\frac{2\alpha}{2\alpha+1}} + \frac{\log p}{n}\right)
\end{equation}

\subsubsection{Lower bound by Assouad's Lemma}
We first establish a lower bound on the minimax risk of estimating $\Omega
\in \Ff_{11}$.  We define the subparameter space as follows.  Given positive
integers $k$ and $m$ such that $2k \leq p$ and $1 \leq m \leq k$, we
parameterize a set of matrices $B(m, k) = (b_{ij})_{p\times p}$ as follows:
\begin{align*}
    b_{ij} = \one\left\{ i = m \text{ and }m+1\leq j \leq 2k, \text{ or }
    j = m \text{ and } m + 1\leq i \leq 2k\right\}
\end{align*}

Let $k = n^\frac{1}{2\alpha + 1}$ and $a = k^{-\alpha - 1}$.  Then, we define
the following set of $2^k$ precision matrices, each parameterized by $\theta
\in \{0, 1\}^k$:
\begin{equation}\label{eq:F11-definition}
    \Ff_{11} = \left\{
    \Omega(\theta): \Omega(\theta) = \II_{p\times p}
    + \tau \alpha \sum_{m=1}^k \theta_m B(m, k)
    \right\}
\end{equation}
with $0 < \tau < 2^{-\alpha - 1}M$.  We may assume without loss of generality
that $M_0 > 1$ and $\rho > 1$.  If that is not the case, we may shrink the
eigenvalues by replacing $\II_{p\times p}$ with $\varepsilon\II_{p\times p}$,
$0 < \varepsilon < \min\{M_0, \rho\}$ as necessary.

We now prove the lower bound in (\ref{eq:F11-lower-bound}).  Suppose $\xx_1,
\dotsc, \xx_n \iid \Nn(0, \Omega(\theta)^{-1})$ with $\Omega \in \Ff_{11}$ and
joint distribution $P_\theta$.  An application of Assouad's Lemma to $\Ff_{11}$
yields the bound:
\begin{equation}\label{eq:applied-assouad}
    \inf_{\hat\Omega}\max_{\theta\in\{0, 1\}^k}
    2^2\EE_\theta\norm{\hat\Omega - \Omega(\theta)}^2
    \geq
    \min_{H(\theta, \theta') \geq 1}
    \frac{\norm{\Omega(\theta) - \Omega(\theta')}^2}{H(\theta, \theta')}
    \cdot
    \frac{k}{2}
    \cdot
    \min_{H(\theta, \theta') = 1}\norm{\PP_\theta \wedge \PP_{\theta'}}
\end{equation}

Lemmas \ref{lemma:assouad-term1} and \ref{lemma:assouad-term3} give
bounds on the first and third terms in Equation
(\ref{eq:applied-assouad}).  The proof of these lemmas may be found in Appendix
\ref{aux-lemma-proofs}.

\begin{lemma}\label{lemma:assouad-term1}
    Let $\Omega(\theta)$ be defined as in Equation~\eqref{eq:F11-definition}.
    Then for some constant $c > 0$
    \begin{align*}
        \min_{H(\theta, \theta') \geq 1}
        \frac{\norm{\Omega(\theta) - \Omega(\theta')}^2}{H(\theta, \theta')}
        \geq cka^2
    \end{align*}
\end{lemma}

\begin{lemma}\label{lemma:assouad-term3}
    Let $\xx_1, \dotsc, \xx_n \iid \Nn(0, \Omega(\theta)^{-1})$ with 
    $\Omega(\theta) \in \Ff_{11}$, with the joint distribution denoted
    by $P_\theta$.  Then:
    \begin{align*}
        \min_{H(\theta, \theta') = 1} \norm{\PP_\theta \wedge \PP_{\theta'}}
        \geq c > 0
    \end{align*}
    for some constant $c > 0$.
\end{lemma}

From Lemmas \ref{lemma:assouad-term1} and \ref{lemma:assouad-term3} and
the fact that $k = n^\frac{1}{2\alpha + 1}$ we may conclude that:
\begin{align*}
    \max_{\Omega(\theta) \in \Ff_{11}}
    2^2 \EE_\theta\norm{\hat\Omega - \Omega(\theta)}^2
    &\geq \frac{c^2}{2}k^2a^2   \\
    &\geq c_1 n^{-\frac{2\alpha}{2\alpha+1}}
\end{align*}
\qed

\subsubsection{Lower bound by Le Cam's Method}
To establish a lower bound on the minimax risk of estimating $\Omega \in
\Ff_{12}$, we first define the subparameter space $\Ff_{12}$ consisting
of diagonal matrices as follows:
\begin{equation}\label{eq:F12-definition}
    \Ff_{12} = \left\{
    \Omega_m = \omega_{ij} = \one\{i = j\}\left(
    1 + \one\{i = j = m\}\sqrt{\frac{\tau}{n}\log p_1}
    \right)^{-1}, 0 \leq m \leq p_1
    \right\}
\end{equation}
where $p_1 = \min\{p, \exp\{\frac{n}{2}\}\}$ and $0 < \tau < \min\{(M_0 - 1)^2,
(\rho - 1)^2, 1\}$.

We establish this lower bound using Le Cam's method.  Denote a set of
distributions $\{\PP_\theta: \theta \in \Theta\}$ where $\Theta = \{\theta_0,
\theta_1, \dotsc, \theta_{p_1}\}$.  Le Cam's method gives a lower bound on the
maximum estimation risk over the parameter set $\Theta$.

Suppose a loss function $L(t, \theta)$ of an estimator $t$ and distribution
parameter $\theta$.  Define $r(\theta_0, \theta_m) = \inf_t\left[ L(t, \theta)
+ L(t, \theta_m)\right]$ and $r_\mathrm{min} = \inf_{1 \leq m \leq p_1} r(\theta_0,
\theta_m)$.  Finally, denote $\bar \PP = \frac{1}{p_1}\sum_{m=1}^{p_1}
\PP_{\theta_m}$.  By Le Cam's method, bounding the total variation affinity
is sufficient to provide a lower bound over the parameter space:
\begin{equation}\label{eq:lecam}
    \sup_\theta L(T, \theta) \geq \frac{1}{2}r_\mathrm{min}\norm{\PP_{\theta_0}
            \wedge \bar\PP}
\end{equation}

We now apply Le Cam's method to the bandable precision matrix estimation
problem.  For $0 \leq m \leq p_1$, let $\Omega_m$ be as defined in $\Ff_{12}$
in Equation (\ref{eq:F12-definition}).  For ease of analysis, we invert each
member of the set $\Ff_{12}$ to create:
\begin{equation}\label{eq:F12'-definition}
    \Ff_{12}' = \left\{
    \Sigma_m: \Sigma_m = \Omega_m^{-1}, \Omega \in \Ff_{12}
    \right\}
\end{equation}
The inversion may be performed trivially as every member of $\Ff_{12}$ is
diagonal.  Then, for $1 \leq i \leq m$, $\Sigma_m$ is a diagonal matrix with:
\begin{align*}
    \sigma_{ii} = \begin{cases}
        1 + \sqrt{\tau\frac{\log p_1}{n}}   \ffor i = m \\
        1 \ffor i \neq m
    \end{cases}
\end{align*}

Suppose we draw $\RR^p \ni \xx_1, \dotsc, \xx_n \iid \Nn(0, \Sigma_m)$, with
joint density $f_m$, $0 \leq m \leq p_1$.  The joint density may be factorized:
\begin{align*}
    f_m = \prod_{1 \leq i \leq n, 1 \leq j \leq p, j \neq m} \phi_1(x_j^i)
    \cdot \prod_{1 \leq i \leq n} \phi_{\sigma_{mm}}(x_m^i)
\end{align*}
where $\phi_\sigma$ denotes the univariate density $\Nn(0, \sigma)$.  From here
on, the proof of the lower bound is identical to that found in Lemma 7 of
\cite{cai_optimal_2010}, but is reproduced here for completeness.

Let $\theta_m = \Sigma_m$ for $0 \leq m \leq p_1$ and the loss function $L$
be the squared operator norm.  First, we establish a bound on
$\norm{\PP_{\theta_0} \wedge \bar \PP}$.  Note that for two arbitrary densities
$q_0, q_1$, we may rewrite the total variation affinity as one minus the total
variation distance:
\[
\int q_0 \wedge q_1 d\mu =  1 - \frac{1}{2}\int |q_0-q_1|d\mu
\]

Then, we may free ourselves of the absolute value by changing the measure
of integration to $q_1$, and then apply Jensen's inequality:
\begin{align}
\left[\int|q_0 - q_1|d\mu\right]^2
    &=  \left[\int\left(\frac{|q_0 - q_1|}{q_1}\right)q_1d\mu\right]^2  \\
    &\leq\int\left(\frac{|q_0 - q_1|}{q_1}\right)^2q_1d\mu \label{line:jensen} \\
    &= \int \frac{q_0^2-2q_0q_1+q_1^2}{q_1}d\mu        \\
    &= \int \frac{q_0^2}{q_1} - 2q_0 + q_1d\mu        \\
    &= \int \frac{q_0^2}{q_1}d\mu - 1
\end{align}

The $q_1^2$ in the denominator allows us to eliminate the $q_1$ outside the
fraction that we treated as our measure of integration when applying Jensen's
inequality in line (\ref{line:jensen}).  This clears a path for us to establish
a bound on the total variation affinity:
\[
\norm{\mathbf{P}_{\theta_0} \wedge \mathbf{\bar P}}
\geq  1 - \frac{1}{2}\left(\int \frac{(\frac{1}{p_1}\sum f_m)^2}{f_0}d\mu - 1
\right)^\frac{1}{2}
\]

That the total variation affinity is bounded away from zero is shown by proving
$\int \frac{(\frac{1}{p_1}\sum f_m)^2} {f_0}d\mu - 1\rightarrow 0$.  We expand
this term:
\begin{align*}
\int \frac{(\frac{1}{p_1}\sum f_m)^2} {f_0}d\mu - 1
    &=  \int \frac{(\frac{1}{p_1}\sum f_m)^2} {f_0}d\mu - 1 \\
    &=  \frac{1}{p_1^2} \int \frac{
            \sum_{m=1}^{p_1} f_m^2 + \sum_{m\neq j} f_m f_j
        } {f_0}d\mu - 1
\end{align*}

\begin{lemma}\label{lemma:lecam-cross-terms}
    For the cross terms \(\frac{f_jf_m}{f_0}\), \(j \neq m\):
    \begin{align*}
        \int \frac{f_jf_m}{f_0} d\mu
        &=  1
    \end{align*}
\end{lemma}

\begin{lemma}\label{lemma:lecam-squared-terms}
    For the squared terms \(\frac{f_m^2}{f_0}\):
    \begin{align*}
        \int \frac{f_m^2}{f_0} d\mu
        &=  \left( 1 - \tau\frac{\log p_1}{n}\right)^{-\frac{n}{2}}
    \end{align*}
\end{lemma}

Let us take $0 < \tau < 1$.  Then we have:
\begin{align*}
\frac{1}{p_1^2}\sum_{m=1}^{p_1}
\left(\frac{f_m^2}{f_0}d\mu - 1\right)
    &\leq \frac{1}{p_1}\left(1 -
        \tau\frac{\log p_1}{n}\right)^{-\frac{n}{2}}
    -   \frac{1}{p_1}   \\
    &=  \exp\left\{-\log p_1 - \frac{n}{2}\log\left(
            1 - \tau\frac{\log p_1}{n}\right)\right\}
        -\frac{1}{p_1}  \\
    &\stackrel{n\rightarrow0}{\longrightarrow} 0
\end{align*}
where we exploit the fact that $\log(1-x)\geq -2x$ for $0 < x <
\frac{1}{2}$.  Combined with the previously proved fact that:
\[
    \int \frac{f_mf_j}{f_0}d\mu -1 = 0
\]

We thus conclude that:
\[
    \frac{1}{p_1^2}\sum_{m=1}^{p_1}\int \frac{f_m^2}{f_0} d\mu
        +
    \frac{1}{p_1^2}\sum_{m \neq j}\int \frac{f_mf_j}{f_0} d\mu
    \rightarrow 0
\]
allowing us to bound:
\[
\norm{\mathbf{P}_{\theta_0} - \mathbf{\bar P}} \geq c
\]

Finally, we give a bound on $r_\mathrm{min}$.  Let $\theta_m = \Sigma_m$ for
$0 \leq m \leq p_1$, and let the loss function $L$ be the squared operator
norm.  Then we see that:
\begin{align*}
r(\theta_0, \theta_m)
    &=  r(\Sigma_0, \Sigma_m)       \\
    &=  \inf_t \left[L(t, \Sigma_0) + L(t, \Sigma_m)\right]
\end{align*}

Observe that the operator norm in $\ell_2$ distance on a diagonal matrix is
simply the largest element.  Then, we may minimize the above quantity with
$t_{ii} = 1 + \frac{1}{2}\sqrt{\tau\frac{\log p_1}{n}}\mathbf{1}\{i = m\}$.  This
gives us:
\begin{align*}
r(\theta_0, \theta_m)
    &=  2\cdot \frac{1}{4}\tau\frac{\log p_1}{n}    \\
    &=  \frac{1}{2}\tau\frac{\log p_1}{n}
\end{align*}
for $1 \leq m \leq p_1$, implying that $r_\mathrm{min} =
\frac{1}{2}\tau\frac{\log p_1}{n}$.  Substituting this result back into
the lower bound given in Equation (\ref{eq:lecam}), we have:
\begin{align*}
\sup_\theta \mathbf{E} L(T, \theta)
    &\geq \frac{1}{2}r_\mathrm{min} \norm{\mathbf{P}_{\theta_0}
    \wedge \mathbf{\bar P}} \\
    &=  \frac{c}{4}\tau\frac{\log p_1}{n}       \\
    &\geq  c\frac{\log p_1}{n}
\end{align*}
where $p_1 = \max\{p, \exp\{\frac{n}{2}\}\}$.
\qed


%% file: simulations.tex
We implemented the blockwise inversion technique in NumPy and ran simulations
on synthetic datasets.  Our experiments confirm that even in the finite sample
case, the blockwise inversion technique achieves the theoretical rates. In the
experiments, we draw observations from a multivariate normal distribution with
precision parameter \(\Omega \in \Ff_\alpha\), as defined in
(\ref{eq:parameter-space}).  Following \cite{cai_optimal_2010}, for given
constants \(\rho, \alpha, p\), we consider precision matrices \(\Omega =
(\omega_{ij})_{1 \leq i, j \leq p}\) of the form:
\begin{equation}\label{eq:simulated-omegas}
    \omega_{ij} = \begin{cases}
        1               \ffor   1 \leq i = j \leq p \\
        \rho|i - j|^{-\alpha - 1} \ffor 1 \leq i \neq j \leq p
    \end{cases}
\end{equation}

Though the precision matrices considered in our experiments are Toeplitz,
our estimator does not take advantage of this knowledge.  We choose
\(\rho = 0.6\) to ensure that the matrices generated are non-negative
definite.

In applying the tapering estimator as defined in (\ref{eq:tapering-estimator}), we
choose the bandwidth to be \(k = \lfloor n^\frac{1}{2\alpha + 1}\rfloor\),
which gives the optimal rate of convergence, as established in Theorem
\ref{thm:upper-bound}.

In our experiments, we varied \(\alpha\), \(n\), and \(p\).  For our first
set of experiments, we allowed \(\alpha\) to take on values in
\(\{0.2, 0.3, 0.4, 0.5\}\), \(n\) to take values in
\(\{250, 500, 750, 1000\}\), and \(p\) to take values in
\(\{100, 200, 300, 400\}\).  Each setting was run for five trials, and the
averages are plotted with error bars to show variability between experiments.  We
observe in Figure \ref{fig:vary-log-p} that the spectral norm error
increases linearly
as \(\log p\) increases, confirming the \(\frac{\log p}{n}\) term in the rate
of convergence.

Building upon the experimental results from the first set of simulations, we
provide an additional sets of trials for the \(\alpha = 0.2, p = 400\) case, with
\(n \in \{11000, 3162, 1670\}\).  These sample sizes were chosen so that in
Figure \ref{fig:vary-n}, there is overlap between the
error plots for
\(\alpha = 0.2\) and the other \(\alpha\) regimes\footnote{
    For the \(\alpha=0.2, p = 400\) case, we omit the settings where
    \(n \in \{250, 500, 750\}\) from Figure \ref{fig:vary-n} to improve the
    clarity of the plot.
}.  As with
Figure \ref{fig:vary-log-p}, Figure \ref{fig:vary-n} confirms the minimax rate
of convergence given in Theorem \ref{thm:upper-bound}. Namely, we see
that plotting the error with respect to $n^{-\frac{2\alpha}{2
    \alpha+1}}$ results in linear plots with almost identical
slopes. We note that in both plots, there is a small
difference in the behavior for the case \(\alpha = 0.2\).
This observation can be attributed to the fact that
for such a slow decay of the precision matrix bandwidth, we have a
more subtle interplay between the bias and variance terms presented in
the theorems above.

\def\imgwidth{0.75}
\begin{figure}
    \centering
    \begin{subfigure}[b]{\imgwidth\textwidth}
        \includegraphics[width=\textwidth]{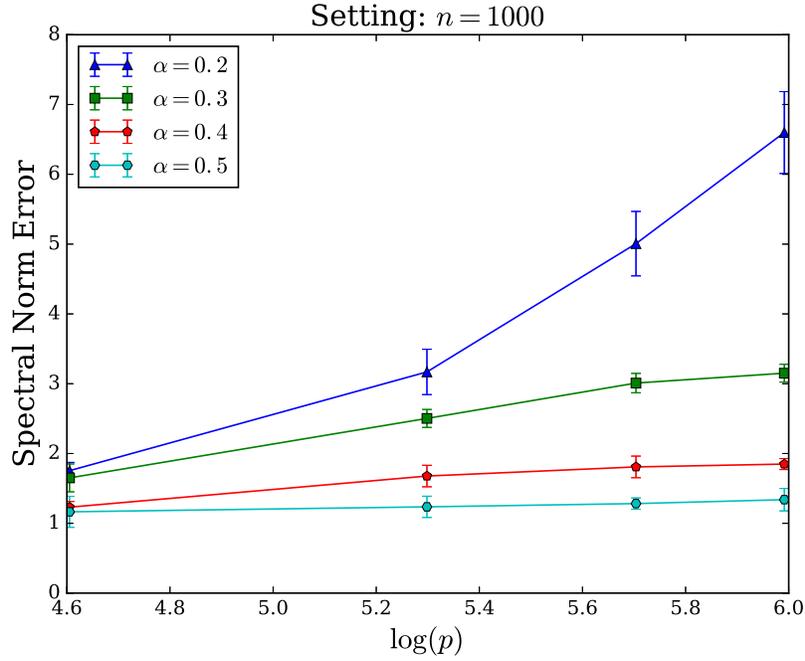}
        \caption{Spectral norm error as \(\log p\) changes.}
        \label{fig:vary-log-p}
    \end{subfigure}
    \begin{subfigure}[b]{\imgwidth\textwidth}
        \includegraphics[width=\textwidth]{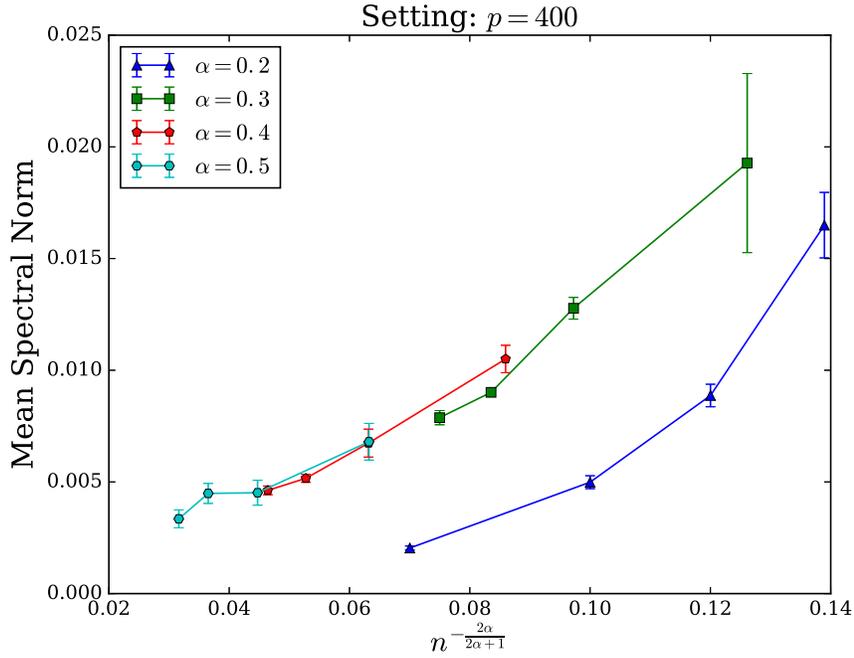}
        \caption{Mean spectral norm error as \(n^{-\frac{2\alpha}{2\alpha+1}}\) changes.}
        \label{fig:vary-n}
    \end{subfigure}
    \caption{Experimental results. Note that the plotted error grows linearly
        as a function of \(\log p\) and \(n^{-\frac{2\alpha}{2\alpha+1}}\),
        respectively, matching the theoretical results; however, the linear
        relationship is less clear in the \(\alpha = 0.2\) case, due to the
        subtle interplay of the error terms.} 
    \label{fig:experiments}
\end{figure}


%% file: discussion.tex
Theorems \ref{thm:upper-bound} and \ref{thm:lower-bound} together establish that
the minimax rate of convergence for estimating precision matrices over the
parameter space \(\Ff_\alpha\) given in Equation (\ref{eq:parameter-space})
is \(n^{-\frac{2\alpha}{2\alpha+1}} + \frac{\log p}{n}\).  The theorems further
imply that the blockwise estimator with \(k = n^\frac{1}{2\alpha+1}\) achieves
this optimal rate of convergence.

As in the bandable covariance case established by~\cite{cai_optimal_2010}, we
may observe that different regimes dictate which term dominates in the rate of
convergence.  In the setting where \(\log p\) is of a lower order than
\(n^\frac{1}{2\alpha+1}\), the \(n^{-\frac{2\alpha}{2\alpha+1}}\) term
dominates, and the rate of convergence is determined by the smoothness parameter
\(\alpha\).  However, when \(\log p\) is much larger than
\(n^\frac{1}{2\alpha+1}\), \(p\) has a much greater influence on the minimax
rate of convergence.

Overall, we have shown how much performance gains can be obtained
through added structural constraints. An interesting line of future
work will be to explore algorithms that uniformly exhibit a smooth
transition between fully banded models and sparse models on the
precision matrix. Such methods could adapt to the structure and allow
for mixtures between banded and sparse precision matrices. The results
presented here apply to the case of subgaussian random
variables. Unfortunately, moving away from the Gaussian setting in
general breaks the connection between precision matrices and graph
structure. Hence, a fruitful line of work will be to also develop
methods that can be applied to general exponential families.


%% file: appendix.tex
\section{Proof of Auxiliary Lemmas}\label{aux-lemma-proofs}
We now present the proofs of the auxiliary lemmas presented above. The
proofs of Lemmas~\ref{lemma:population-sum-of-blocks}
and~\ref{lemma:bound-by-max} are closely related to the analogous
proofs of~\cite{cai_optimal_2010}. For completeness, we adapt those
proofs below.
\subsection{Proof of Lemma \ref{lemma:population-sum-of-blocks}}
Consider a pair \(i, j\), indexing an entry of \(\Omega_A\).  Without loss of
generality, assume that \(i \leq j\).  Then, for a fixed \(l, m\), the set
\(\{i, j\}\) is contained in \(\{l, \dotsc, l + m - 1\}\) if and only if
\(l \leq i \leq j \leq l + m - 1\).  This condition is equivalently stated
as \(j - m + 1 \leq l \leq i\).

Keeping \(i, j, m\) fixed, \(\Card\{l: j - m + 1 \leq l \leq i\}\) intuitively
gives the number of \(\Omega_l^{(m)}\) that are nonzero on the  
\(ij^\mathrm{th}\) entry.  Observe that:
\begin{align*}
    \Card\{l: j - m + 1 \leq l \leq i\}
        &=  \left( i - (j - m + 1) + 1\right)_+  \\
        &=  \left( m - |i - j|\right)_+
\end{align*}
It immediately follows that:
\begin{align*}
    \Card\{l: j - k + 1 \leq l \leq i\}
        &=  \left( k - |i - j|\right)_+ \\
    \Card\{l: j - \nicefrac{k}{2} + 1 \leq l \leq i\}
        &=  \left( \nicefrac{k}{2} - |i - j|\right)_+ \\
\end{align*}
Therefore, the tapering coefficient may be expressed:
\begin{align*}
    \frac{k}{2}\cdot\TaperCoef
        &=  \left(k - |i - j|\right)_+
          - \left(\nicefrac{k}{2} - |i - j|\right)_+  \\
        &=  \Card\{l: j - k + 1 \leq l \leq i\}
          - \Card\{l: j - \nicefrac{k}{2} + 1 \leq l \leq i\}
\end{align*}
\qed

\subsection{Proof of Lemma \ref{lemma:bound-by-max}}
By construction, we may decompose:
\begin{equation}\label{eq:decompose-spectral-diff}
    \norm{\hat\Omega_m - \Omega_A}
    \leq
    \frac{2}{m}\cdot\left(
        \norm{\sum_{l=1-m}^p \hat\Omega_l^{(m)} - \Omega_l^{(m)}}
        -
        \norm{\sum_{l=1-\nicefrac{m}{2}}^p \hat\Omega_l^{(\nicefrac{m}{2})} - \Omega_l^{(\nicefrac{m}{2})}}
    \right)
\end{equation}
Assume without loss of generality that \(p\) is divisible by \(m\).  We may
rewrite:
\begin{align*}
    \norm{\sum_{l=1-m}^p \hat\Omega_l^{(m)} - \Omega_l^{(m)}}
    &=  \norm{\sum_{j=-1}^{\nicefrac{p}{m}}\sum_{l=1}^m
          \hat\Omega_{jm+l}^{(m)} - \Omega_{jm+l}^{(m)}} \\
    &\leq
        \sum_{l=1}^m\norm{\sum_{j=-1}^{\nicefrac{p}{m}}
        \hat\Omega_{jm+l}^{(m)} - \Omega_{jm+l}^{(m)}} \\
    &\leq
        m\cdot\max_{1\leq l\leq m}\norm{\sum_{j=-1}^{\nicefrac{p}{m}}
        \hat\Omega_{jm+l}^{(m)} - \Omega_{jm+l}^{(m)}}
\end{align*}
Because \(\hat\Omega_{jm+l}^{(m)},\Omega_{jm+l}^{(m)}\) are disjoint diagonal
subblocks over \(-1 \leq j \leq \nicefrac{p}{m}\), it follows that:
\begin{equation}\label{eq:bound-by-max}
    \norm{\sum_{l=1-m}^p \hat\Omega_l^{(m)} - \Omega_l^{(m)}}
    \leq
        m\cdot\max_{1-m\leq l\leq p}\norm{
        \hat\Omega_{jm+l}^{(m)} - \Omega_{jm+l}^{(m)}} \\
\end{equation}
Therefore, from Equations (\ref{eq:decompose-spectral-diff}) and
(\ref{eq:bound-by-max}) we have:
\begin{align*}
    \norm{\hat\Omega_m - \Omega_A}
    &\leq
        2 N^{(m)} + N^{(\nicefrac{m}{2})} \\
    &\leq
        C\cdot N^{(m)}
\end{align*}

\subsection{Proof of Lemma \ref{lemma:schur-correction}}
Assume without loss of generality that \(l \in [2m+1, p-3m]\).  This argument
can be extended to the periphery easily.

Consider a given \(m \times m\) block of interest \(\PrecCenterBlock\).  We
denote the \(3m \times 3m\) block centered about \(\PrecCenterBlock\) by
\(\PrecPaddedBlock\).  We permute the rows and
columns of \(\Omega\) such that \(\PrecPaddedBlock\) is located in the upper
left.  This permuted matrix may be expressed as:
\begin{align*}
    \pi(\Omega)
    &=  \bmat{
            \Omega_{l-m \leq i, j \leq p}  
        &   \Omega_{l-m\leq i \leq p, 1 \leq j < l-m}  \\
            \Omega_{l-m\leq i \leq p, 1 \leq j < l-m}^\top
        &   \Omega_{1 \leq i, j < l - m}
    }   \\
    &=  \bmat{
        \Omega_{l-m}^{(3m)} &   \Omega_B    \\
        \Omega_B^\top       &   \Omega_C
    }
\end{align*}
Let us express:
\begin{align*}
    \Omega_B = \bmat{\Omega_{B_1} \\ \Omega_{B_2} \\ \Omega_{B_3}}
\end{align*}
with \(\Omega_{B_2}\) spanning the indices:
\begin{align*}
    \Omega_{B_2} = \bmat{
        \Omega_{l \leq i < l+m, l + 2m \leq j \leq p}
        &
        \Omega_{l \leq i < l+m, 1 \leq j < l - m}
    }
\end{align*}
By examining indices, we see that \(\Omega_{B_2}\) has no entries within \(m\)
of the diagonal.  

Now, consider \(\Sigma = \pi(\Omega)^{-1}\).  If we take the upper left \(3m
    \times 3m\) block \(\Sigma_{3m}\) of \(\Sigma\) and invert it by the Schur
complement, we obtain:
\begin{align*}
    \Sigma_{3m}^{-1}
    &=  \Omega_{l - m}^{(3m)} - \Omega_B \Omega_C^{-1} \Omega_B^\top
\end{align*}
At this point, we note that the central \(m \times m\) block of
\(\Omega_{l-m}^{(3m)}\) is in fact \(\Omega_l^{(m)}\), and denote the central
\(m \times m\) block of \(\Sigma_{3m}^{-1}\) by \(\tilde\Omega_l^{(m)}\).  
Therefore, we may write:
\begin{align*}
    \tilde\Omega_l^{(m)}
    &= \Omega_l^{(m)} - \Omega_{B_2}\Omega_C^{-1}\Omega_{B_2}^\top
\end{align*}
\qed

\subsection{Proof of Lemma \ref{lemma:correction-rate}}
Recall from Lemma \ref{lemma:schur-correction} that
\(W = \Omega_{B_2}\Omega_C^{-1}\Omega_{B_2}^\top\), where \(\Omega_{B_2}\)
has no in-band entries.  First, we bound, with \(k = m\):
\begin{align*}
    \norm{\Omega_{B_2}}
    &\leq \norm{\bmat{0 & \Omega_{B_2} \\ \Omega_{B_2}^\top & 0}}   \\
    &\leq \norm{\Omega_{B_2}}_1 \\
    &\leq Mk^{-\alpha}
\end{align*}
Next, we bound the spectral norm of \(\Omega_C^{-1}\).  Note that this is
equivalent to bounding \(\lambda_\mathrm{min}(\Omega_C)\) away from zero.
\begin{align*}
    \lambda_\mathrm{min}(\Omega_C)
    &=  \min_{\substack{\tilde \vv \in \RR^{p - 3m},\\ \norm{\tilde \vv} = 1}}
        \norm{\Omega_C\tilde \vv} \\
    &=  \min_{\substack{\vv \in \RR^p, \\ \norm{\vv} = 1,
            \\ \vv_{\supp(C)^c} = 0}}
        \norm{\Omega \vv} \\
    &\geq \min_{\substack{\vv \in \RR^p,\\ \norm{\vv} = 1}}
        \norm{\Omega \vv} \\
    &\geq \frac{1}{M_0}
\end{align*}
Therefore, we may conclude:
\begin{align*}
    \norm{W}
    &\leq \norm{\Omega_{B_2}}^2\norm{\Omega_C^{-1}} \\
    &\leq Cm^{-2\alpha}
\end{align*}
\qed

\subsection{Proof of Lemma \ref{lemma:max-conc-bound}}
Note that for arbitrary \(l, m\), we have:
\begin{align*}
    \norm{\hat\Omega_l^{(m)} - \tilde\Omega_l^{(m)}}
    &\leq 
    \norm{\hat\Omega_{l-m}^{(3m)} - \tilde\Omega_{l-m}^{(3m)}}  \\
    &=  \norm{\hat\Sigma_{3m}^{-1} - \Sigma_{3m}^{-1}}  \\
    &\leq  \norm{\Sigma_{3m}^{-1}}\norm{\Sigma_{3m} - \hat\Sigma_{3m}}
        \norm{\hat\Sigma_{3m}^{-1}}  \\
\end{align*}

We now consider the spectral norm of the matrix
\(\hat\Sigma_{3m}^{-1}\).
We may bound the minimum eigenvalue of \(\hat\Sigma_{3m}\)
away from zero with high probability.

By decomposing
\(\hat\Sigma_{3m} = \Sigma_{3m} + (\hat\Sigma_{3m} - \Sigma_{3m})\) and
applying Weyl's Theorem \citep{horn_matrix_2012}:
\begin{align*}
    \lambda_\mathrm{min}(\hat\Sigma_{3m})
    &\geq   \lambda_\mathrm{min}(\Sigma_{3m})
            + \lambda_\mathrm{min}(\hat\Sigma_{3m} - \Sigma_{3m})   \\
    &\geq   \lambda_\mathrm{min}(\Sigma_{3m})
            - \norm{\hat\Sigma_{3m} - \Sigma_{3m}}   \\
\end{align*}

We now state a useful result for bounding the spectral norm of a random matrix.
\begin{lemma}\label{lemma:ep-net-arg}
    Suppose \(A\) is an \(m \times n\) subgaussian random matrix.  Then
    there exists some \(\rho > 0\) such that:
    \begin{equation}\label{eq:subgaussian-spectral-bound}
        \PP\left\{
            \norm{A} > t
        \right\}
        \leq
        5^{m+n} \exp\left\{
            -t^2\rho
        \right\}
    \end{equation}
    Furthermore, let \(\xx_1, \dotsc, \xx_n \in \RR^m\) be i.i.d. vectors with
    \(\EE (\xx_i - \mu)(\xx_i - \mu)^\top = \Sigma\).  Denote the empirical
    covariance matrix \(\hat\Sigma = \frac{1}{n}\sum_{i=1}^n (\xx_i -
    \xbar)(\xx_i - \xbar )^\top\).  Then for some \(\tilde\rho > 0\):
    \begin{equation}\label{eq:subgaus-sigma-spectral-bound}
        \PP\left\{
            \norm{\hat\Sigma - \Sigma} > x
        \right\}
        \leq
        {25}^m \exp\left\{
            - \frac{ nx^2\tilde\rho}{2}
        \right\} + 5^m \exp\left\{
            - \frac{n x \tilde\rho }{2} 
        \right\}
    \end{equation}
    for all \(0 < x < \tilde\rho\).
\end{lemma}
By the above result there exists some
\(\tilde\rho > 0\) such that:
\begin{align*}
    \PP\left\{
        \norm{\Sigma_{3m} - \hat\Sigma_{3m}} > x
    \right\}
    &\leq
      {25}^{3m}\exp\left\{
        -\frac{nx^2\tilde\rho}{2}
    \right\} \, + \, 5^{3m} \exp\left\{
        -\frac{n x \tilde\rho}{2}
    \right\}
\end{align*}
for all \(0 < x < \tilde \rho\).
Choose \(x = 2\sqrt\frac{m + \log p}{n\tilde\rho}\) and note that $x =
o(1)$, by our assumptions. Thus,
\begin{align*}
    \PP\left\{
        \norm{\Sigma_{3m} - \hat\Sigma_{3m}}
        > 2\sqrt\frac{m + \log p}{n\tilde\rho}
    \right\}
    &= O\left(p^{-4}\right)
\end{align*}
Assume \(n\rho_1 > 4 M_0^2 (m + \log p)\).  Then, with high probability, we
have: 
\begin{align*}
    \lambda_\mathrm{min}(\hat\Sigma_{3m})
    &\geq   \lambda_\mathrm{min}(\Sigma_{3m}) - 2\cdot \frac{1}{2M_0} \\
    &=      \lambda_\mathrm{min}(\Sigma_{3m}) - \frac{1}{2M_0} \\
    &=      \lambda_\mathrm{min}(\Sigma_{3m})
            - \frac{1}{2} \lambda_\mathrm{min}(\Sigma_{3m})\\
    &=      \frac{1}{2}M_0  >      c > 0
\end{align*}
Therefore, it follows that:
\begin{align*}
    \norm{\hat\Omega_l^{(m)} - \tilde\Omega_l^{(m)}}
    &\leq C\norm{\Sigma_{3m} - \hat\Sigma_{3m}}
\end{align*}
with high probability.  We now re-apply the concentration bound from Lemma
\ref{lemma:ep-net-arg}.  There exists a constant \(\rho_1 > 0\) such that:
\begin{align*}
    \PP\left\{
        \norm{\hat\Omega_l^{(m)} - \tilde\Omega_l^{(m)}} > Cx
    \right\}
    &=  
    \PP\left\{
        \norm{\Sigma_{3m} - \hat\Sigma_{3m}} > x
    \right\}    \\
    &\leq
      {25}^{3m}\exp\left\{
        -nx^2\rho_1
    \right\}
\end{align*}
Then, by the union bound, we have:
\begin{align*}
    \PP\left\{
        \max_{1 \leq l \leq p - m + 1}
        \norm{\hat\Omega_l^{(m)} - \tilde\Omega_l^{(m)}} > Cx
    \right\}
    &\leq
    \sum_{1 \leq l \leq p - m + 1}
    \PP\left\{
        \norm{\hat\Omega_l^{(m)} - \tilde\Omega_l^{(m)}} > Cx
    \right\}    \\
    &\leq
    2p\cdot {25}^{3m}\exp\left\{
        -nx^2\rho_1
    \right\}
\end{align*}
\qed

\subsection{Proof of Lemma \ref{lemma:assouad-term1}}
Let $\Omega(\theta) \in \Ff_{11}$ be defined as in Equation
(\ref{eq:F11-definition}).  We wish to show that:
\begin{align*}
    \min_{H(\theta, \theta') \geq 1}
    \frac{\norm{\Omega(\theta) - \Omega(\theta')}^2}{H(\theta, \theta')}
    \geq cka^2
\end{align*}

Define $v = (\one\{k \leq i \leq 2k\})_i \in \RR^p$, $w = [\Omega(\theta)
- \Omega(\theta')]v$.  Observe that there are exactly $H(\theta, \theta')$
entries in $w$ such that $|w_i| = \tau ka$.  Further note that $\norm{v}_2^2 
= k$.  This implies:
\begin{align*}
    \norm{\Omega(\theta) - \Omega(\theta')}^2
    &\geq \frac{\norm{[\Omega(\theta) - \Omega(\theta')]v}_2^2}{\norm{v}_2^2}   \\
    &\geq \frac{H(\theta, \theta')\cdot(\tau ka)^2}{k}  \\
    &=  H(\theta, \theta') \tau^2 ka^2
\end{align*}
It then follows that:
\begin{align*}
    \frac{\norm{\Omega(\theta) - \Omega(\theta')}^2}{H(\theta, \theta')}
    &\geq \tau^2ka^2    \\
    \Rightarrow
    \min_{H(\theta, \theta') \geq 1}
    \frac{\norm{\Omega(\theta) - \Omega(\theta')}^2}{H(\theta, \theta')}
    &\geq cka^2
\end{align*}
\qed

\subsection{Proof of Lemma \ref{lemma:assouad-term3}}
Let $\xx_1, \dotsc, \xx_n \iid \Nn(\Omega(\theta))$ with $\Omega(\theta)
\in \Ff_{11}$ as defined in Equation (\ref{eq:F11-definition}), with joint
distribution $\PP_\theta$.  We wish to show that for some constant $c > 0$:
\begin{align*}
    \min_{H(\theta, \theta')=1}\norm{\PP_\theta \wedge \PP_{\theta'}} \geq c
\end{align*}
Because $\norm{\PP_1 \wedge \PP_2} = 1 - \frac{1}{2} \norm{\PP_1 - \PP_2}_1$,
it is sufficient to show that:
\begin{align*}
    \norm{\PP_\theta - \PP_{\theta'}}_1^2 \leq c
\end{align*}
We may bound the squared $\ell_1$-norm from above by the Kullback-Leibler
Divergence:
\begin{align*}
    \norm{\PP_\theta - \PP_{\theta'}}_1^2
    &\leq 2 D_\mathrm{KL}(\PP_\theta | \PP_{\theta'})   \\
    &=  2n\left[
    \frac{1}{2} \tr\left(\Omega^{-1}(\theta')\Omega(\theta)\right)
    - \frac{1}{2}\log\det\left(\Omega^{-1}(\theta')\Omega(\theta)\right)
    - \frac{p}{2}
    \right]
\end{align*}
This reduced form of the Kullback-Leibler Divergence is a consequence of the
zero mean for the distribution of $\PP_\theta, \PP_{\theta'}$.  We will show
that we may bound this quantity from above by a constant.

First, we define $D_1 \defas \Omega(\theta) - \Omega(\theta')$.  Observe that:
\begin{align*}
    \Omega^{-1}(\theta')D_1     &=  \Omega^{-1}(\theta')\Omega(\theta) - \II_p  \\
    \Leftrightarrow
    \Omega^{-1}\Omega(\theta)   &=  \Omega^{-1}(\theta')D_1 + \II_p
\end{align*}
From these identities, we may rewrite:
\begin{align*}
    \frac{1}{2}\tr\left(\Omega^{-1}(\theta')\Omega(\theta)\right) - \frac{p}{2}
    &=  \frac{1}{2}\tr\left(\Omega^{-1}(\theta')D_1\right)
\end{align*}
Denote the eigenvalues of $\Omega^{-1}(\theta')D_1$ by $\lambda_i$.  Then:
\begin{align*}
    \tr\left(\Omega^{-1}(\theta')D_1\right)    &=  \sum_{i=1}^p \lambda_i
\end{align*}
We may then bound the spectrum of $\Omega^{-1}(\theta')D_1$ by bounding
the spectrum of the similar matrix $\Omega^{-\frac{1}{2}}(\theta')D_1
\Omega^{-\frac{1}{2}}(\theta')$:
\begin{align*}
    \norm{\Omega^{-1}(\theta')D_1}
    &=  \norm{\Omega^{-\frac{1}{2}}(\theta')D_1\Omega^{-\frac{1}{2}}(\theta')}  \\
    &\leq   \norm{\Omega^{-\frac{1}{2}}}\norm{D_1}\norm{\Omega^{-\frac{1}{2}}}  \\
    &\leq   c_1 \norm{D_1}  \\
    &\leq   c_1 \norm{D_1}_1    \\
    &\leq   c_2 ka
\end{align*}
where $\norm{A}_1 = \max_j \norm{A_{.j}}_1$ denotes the matrix $\ell_1$ norm.
This bound on the spectrum implies that $\lambda_i \in [-c_2ka, c_2ka]$, with
$ka = k^{-\alpha} = n^{-\frac{\alpha}{2\alpha+1}} \rightarrow 0$.  

Now, we bound the $\log\det\left(\Omega^{-1}(\theta')\Omega(\theta)\right)$ term:
\begin{align*}
    \log\det\left(\Omega^{-1}(\theta')\Omega(\theta)\right)
    &=  \log\det\left(\II_p + \Omega^{-1}(\theta') D_1\right)  \\
    &=  \sum_{i=1}^p \log(1 + \lambda_i)    \\
    &=  \sum_{i=1}^p \lambda_i + \left(\log(1 + \lambda_i) - \lambda_i\right)  \\
    &\geq  \tr\left(\Omega^{-1}(\theta')D_1\right)
        + \sum_{i=1}^p \left[\frac{\lambda_i}{1 + \lambda_i}
        - \frac{\lambda_i + \lambda_i^2}{1 + \lambda_i}\right]  \\
    &=  \tr\left(\Omega^{-1}(\theta')D_1\right)
        - \sum_{i=1}^p \frac{\lambda_i^2}{1 + \lambda_i}    \\
    &\geq  \tr\left(\Omega^{-1}(\theta')D_1\right)
        - c_3 \sum_{i=1}^p \lambda_i^2
\end{align*}
where we may bound:
\begin{align*}
    \sum_{i=1}^p \lambda_i^2
        &=  \norm{\Omega^{-1}(\theta')D_1}_F^2  \\
        &=  \norm{\Omega^{-\frac{1}{2}}(\theta')
            D_1\Omega^{-\frac{1}{2}}(\theta')}_F^2  \\
        &\leq  \norm{\Omega^{-\frac{1}{2}}(\theta')}^2
            \norm{D_1}_F^2
            \norm{\Omega^{-\frac{1}{2}}(\theta')}^2 \\
        &\leq c_4ka^2
\end{align*}
due to $H(\theta, \theta') = 1$.  This in turn implies that:
\begin{align*}
    \log\det\left(\Omega^{-1}(\theta')\Omega(\theta)\right)
    &\geq  \tr\left(\Omega^{-1}(\theta')D_1\right)
        - c_3 \sum_{i=1}^p \lambda_i^2  \\
    &\geq  \tr\left(\Omega^{-1}(\theta')D_1\right)
        - c_5 ka^2  \\
    \Rightarrow
    -\frac{1}{2} \log\det\left(\Omega^{-1}(\theta')\Omega(\theta)\right)
    &\leq - \frac{1}{2}\tr\left(\Omega^{-1}(\theta')D_1\right) + \frac{c_5}{2}ka^2
\end{align*}
Finally, this results in the bound:
\begin{align*}
    \norm{\PP_\theta - \PP_{\theta'}}_1^2
    &\leq 2n\left[
    \frac{1}{2}\tr\left(\Omega^{-1}(\theta')D_1\right)
    -\frac{1}{2}\log\det\left(\Omega^{-1}(\theta')\Omega(\theta)\right)
    \right] \\
    &\leq 2n \cdot \frac{c_5}{2} ka^2   \\
    &=  c_5nka^2  \\
    &=  c_5nk^{-2\alpha - 1}  \\
    &=  c_5 \cdot n \cdot n^{-\frac{2\alpha + 1}{2\alpha + 1}}  \\
    &=  c_5
\end{align*}
This immediately implies that:
\begin{align*}
    \norm{\PP_\theta \wedge \PP_{\theta'}} \geq c > 0
\end{align*}
\qed

\subsection{Proof of Lemma \ref{lemma:lecam-cross-terms}}
We can directly evaluate, for all $j, m$:
\begin{align*}
\int \frac{f_jf_m}{f_0} d\mu
&=  \int
    \frac{
        \prod_{\substack{1\leq i \leq n\\1 \leq k \leq p_1\\k\neq j}}
        \phi_1(x_k^i)
        \prod_{\substack{1\leq i \leq n\\1 \leq k \leq p_1\\k\neq m}}
        \phi_1(x_k^i)
        \prod_{1 \leq i \leq n}
        \phi_{\sigma_{mm}}(x_m^i)\phi_{\sigma_{mm}}(x_j^i)
    }{
        \prod_{\substack{1 \leq i \leq n\\1\leq k \leq p_1}} \phi_1(x_k^i)
    }d\left\{x_k^i\right\}_{\substack{1 \leq i \leq n\\1\leq k \leq p_1}}\\
&\text{(Independence.)} \\
&=  \prod_{1\leq i \leq n}\int
    \frac{
        \left[\prod_{\substack{1 \leq k \leq p_1\\k\neq j}}
        \phi_1(x_k^i)\right]
        \left[\prod_{\substack{1 \leq k \leq p_1\\k\neq m}}
        \phi_1(x_k^i)\right]
        \phi_{\sigma_{mm}}(x_m^i)\phi_{\sigma_{mm}}(x_j^i)
    }{
        \prod_{1\leq k \leq p_1} \phi_1(x_k^i)
    }d\left\{x_k^i\right\}_{1\leq k \leq p_1}   \\
&=  \prod_{1\leq i\leq n}\int
    \left[\prod_{\substack{1 \leq k \leq p_1\\k \not\in\{j, m\}}}
    \phi_1(x_k^i)\right]
    \phi_{\sigma_{mm}}(x_m^i)\phi_{\sigma_{mm}}(x_j^i)
    d\left\{x_k^i\right\}_{1\leq k \leq p_1}    \\
&\text{(Independence.)} \\
&=  \prod_{1\leq i\leq n}
    \left[
    \left[
    \prod_{\substack{1 \leq k \leq p_1\\k \not\in\{j, m\}}}
    \int
    \phi_1(x_k^i)d x_k^i
    \right]
    \int\phi_{\sigma_{mm}}(x_m^i)d x_m^i
    \int\phi_{\sigma_{mm}}(x_j^i)d x_j^i
    \right] \\
&=  \prod_{1\leq i\leq n}
    1   \\
&=  1
\end{align*}
\qed

\subsection{Proof of Lemma \ref{lemma:lecam-squared-terms}}
For the squared terms:
\begin{align*}
\int \frac{f_m^2}{f_0} d\mu
&=  \int
    \frac{
        \prod_{\substack{1\leq i \leq n\\1 \leq j \leq p_1\\j\neq m}}
        \phi_1(x_j^i)^2
        \prod_{1 \leq i \leq n}
        \phi_{\sigma_{mm}}(x_m^i)^2
    }{
        \prod_{\substack{1 \leq i \leq n\\1\leq j \leq p_1}} \phi_1(x_j^i)
    }d\left\{x_j^i\right\}_{\substack{1 \leq i \leq n\\1\leq j \leq p_1}}\\
&\text{(Independence.)} \\
&=  \prod_{1\leq i \leq n}\int
    \frac{
        \left[\prod_{\substack{1 \leq j \leq p_1\\j\neq m}}
        \phi_1(x_j^i)^2\right]
        \phi_{\sigma_{mm}}(x_m^i)^2
    }{
        \prod_{1\leq j \leq p_1} \phi_1(x_j^i)
    }d\left\{x_j^i\right\}_{1\leq j \leq p_1}   \\
&=  \prod_{1\leq i \leq n}\int
    \frac{
        \left[\prod_{\substack{1 \leq j \leq p_1\\j\neq m}}
        \phi_1(x_j^i)\right]
        \phi_{\sigma_{mm}}(x_m^i)^2
    }{
        \phi_1(x_m^i)
    }d\left\{x_j^i\right\}_{1\leq j \leq p_1}   \\
&=  \prod_{1\leq i \leq n}
    \left[
    \prod_{\substack{1 \leq j \leq p_1\\j\neq m}}
    \int
    \phi_1(x_j^i)dx_j^i
    \right]
    \int
    \frac{
        \phi_{\sigma_{mm}}(x_m^i)^2
    }{
        \phi_1(x_m^i)
    }dx_m^i   \\
&=  \prod_{1\leq i \leq n}
    \int
    \frac{
        \phi_{\sigma_{mm}}(x_m^i)^2
    }{
        \phi_1(x_m^i)
    }dx_m^i   \\
\end{align*}
We now substitute in the form of the density functions and move the
normalization terms out of the integral:
\begin{align*}
\prod_{1\leq i \leq n}
    \int
    \frac{
        \phi_{\sigma_{mm}}(x_m^i)^2
    }{
        \phi_1(x_m^i)
    }dx_m^i
&=  \frac{
        (\sqrt{2\pi\sigma_{mm}})^{-2n}
    }{
        (\sqrt{2\pi})^{-n}
    }
    \prod_{1\leq i \leq n}
    \int
    \exp\left\{-2\cdot\frac{(x_m^i)^2}{2\sigma_{mm}}\right\}
    \exp\left\{\frac{(x_m^i)^2}{2}\right\}\\
&=  \frac{
        (\sqrt{2\pi\sigma_{mm}})^{-2n}
    }{
        (\sqrt{2\pi})^{-n}
    }
    \prod_{1\leq i \leq n}
    \int
    \exp\left\{
    -\frac{1}{2}\left[
    \frac{(x_m^i)^2}{\frac{\sigma_{mm}}{2 - \sigma_{mm}}}
    \right]
    \right\}    \\
&=  \frac{
        (\sqrt{2\pi\sigma_{mm}})^{-2n}
    }{
        (\sqrt{2\pi})^{-n}
    }
    \left(
    \frac{2\pi\sigma_{mm}}{2 - \sigma_{mm}}
    \right)^\frac{n}{2} \\
&=  (\sqrt{\sigma_{mm}})^{-2n}
    \left(
    \frac{\sigma_{mm}}{2 - \sigma_{mm}}
    \right)^\frac{n}{2} \\
&=  (\sqrt{\sigma_{mm}})^{-n}(\sqrt{2 - \sigma_{mm}})^{-n}  \\
&=  \left[
    2\sigma_{mm} - \sigma_{mm}^2
    \right]^{-\frac{n}{2}}  \\
&=  \left[
    1 - (1 - \sigma_{mm})^2
    \right]^{-\frac{n}{2}}  \\
&=  \left( 1 - \tau\frac{\log p_1}{n}\right)^{-\frac{n}{2}}
\end{align*}
\qed

\subsection{Proof of Lemma \ref{lemma:ep-net-arg}}
We proceed with an $\epsilon$-net
argument. The proof is done in the case $m=n$,
and the extension to the general case is immediate. Let \(S^{m-1}\)
denote the \(\ell_2\)-sphere
in \(\RR^m\),
and let \(S_{\nicefrac{1}{2}}^{m-1}\)
be a \(\nicefrac{1}{2}\)-net
of \(S^{m-1}\).
Note that for every \(\uu \in S^{m-1}\),
there exists \(\xx \in S_{\nicefrac{1}{2}}^{m-1}\),
\(\vv \in \RR^m : \norm{v} \leq \nicefrac{1}{2}\)
such that \(\uu = \xx + \vv\).
Then, for any matrix \(A \in \RR^{m \times m}\),
we may discretize the sphere \(S^{m-1}\):
\begin{align*}
    \norm{A}
    &=  \sup_{\uu \in S^{m-1}} \norm{A\uu}   \\
    &\leq
    \sup_{\xx \in S_{\nicefrac{1}{2}}^{m-1}} \norm{A\xx}
    + \sup_{\vv: \norm{\vv} \leq \frac{1}{2}} \norm{A\vv} \\
    &\leq
    \sup_{\xx \in S_{\nicefrac{1}{2}}^{m-1}} \norm{A\xx}
    + \frac{1}{2}\norm{A}   \\
    \Rightarrow
    \norm{A}
    &\leq
    2\sup_{\xx \in S_{\nicefrac{1}{2}}^{m-1}} \norm{A\xx}
\end{align*}
Following a similar line of reasoning:
\begin{align*}
    \norm{A\xx}
    &=  \sup_{\yy \in S^{m-1}} \langle A\xx, \yy \rangle  \\
    &\leq   2\sup_{\yy \in S_{\nicefrac{1}{2}}^{m-1}} \langle A\xx, \yy\rangle    \\
\end{align*}
Then, by symmetry, we have:
\begin{align*}
    \norm{A}
  & \leq 4 \sup_{\xx, \yy \in S_{\nicefrac{1}{2}}^{m-1}}
        |\xx^\top A \yy|    \\
\end{align*}
From packing arguments, we have that the cardinality of
\(S_{\nicefrac{1}{2}}^{m-1}\) is at most \(5^m\).  Therefore, there exist
\(\vv_1, \dotsc, \vv_{5^m} \in S^{m-1}\)
such that for all
\(A \in \RR^{m \times m}\),
\begin{align*}
    \norm{A} \leq 4 \max_{i,j \leq 5^m} |\vv_i^\top A \vv_j|
\end{align*}
Recall that all entries of the matrix \(A\) are subgaussian with
mean zero.  Now, we note each entry of the vector \(A\vv_j\) is the result of
an inner product between \(A_{i.}\) and a unit vector \(\vv_j\in
S_{\nicefrac{1}{2}}^{m-1} \subset S^{m-1}\); therefore, the entries of
\(A\vv_j\) are also subgaussian distributed.  We repeat this argument to
note that \(\vv_j^\top A\vv_j\) is subgaussian distributed for all
\(i, j\).
Therefore, we may observe that:
\begin{align*}
    \PP\left\{
        \norm{A} \geq t
    \right\}
    &\leq
    \PP\left\{
        \max_{i,j \leq 5^m} |\vv_i^\top A \vv_j| \geq \frac{t}{4}
    \right\}    \\
    &\leq
    \sum_{i,j \leq 5^m} \PP\left\{
        |\vv_i^\top A \vv_j| \geq \frac{t}{4}
    \right\}    \\
    &\leq
      {25}^m\max_{i,j \leq 5^m}
    \PP\left\{
        |\vv_j^\top A \vv_j| \geq \frac{t}{4}
    \right\}    \\
    &\leq {25}^m \exp\left\{
        -t^2\rho
    \right\}
\end{align*}
for some \(\rho > 0\), by the definition of subgaussianity.  This proves
(\ref{eq:subgaussian-spectral-bound}).  We now apply this result to show
(\ref{eq:subgaus-sigma-spectral-bound}).

Recall that we have drawn
\(\xx_1, \dotsc, \xx_n \in \RR^m\) from a subgaussian distribution with
population covariance \(\Sigma\). We wish to bound $\norm{\hat\Sigma -
  \Sigma}$ where $\hat \Sigma = \frac{1}{n} \sum_{i=1}^n (\xx_i - \xbar)(\xx_i -
\xbar)^\top$. We may subtract the mean $\mu$ from both terms, yielding
$\hat \Sigma = \frac{1}{n} \sum_{i=1}^n ((\xx_i - \mu) - (\xbar-\mu))((\xx_i-\mu) -
(\xbar-\mu))^\top$, which can be further simplified to
\begin{equation*}
  \hat \Sigma = \frac{1}{n} \sum_{i=1}^n (\xx_i - \mu)(\xx_i-\mu)^\top - (\xbar-\mu) (\xbar-\mu)^\top
\end{equation*}
Hence,
\begin{align*}
  \norm{\hat\Sigma - \Sigma} & \leq \norm{\frac{1}{n} \sum_{i=1}^n
    (\xx_i - \mu)(\xx_i-\mu)^\top - \Sigma} +
                               \norm{(\xbar-\mu)(\xbar-\mu)^\top} \\
                             & = \norm{\frac{1}{n} \sum_{i=1}^n
    (\xx_i - \mu)(\xx_i-\mu)^\top - \Sigma} +
                               \|\xbar-\mu\|_2^2
\end{align*}
Thus,
\begin{equation*}
  \PP \left \{\norm{\hat \Sigma - \Sigma} \geq x \right \}\leq
  \PP \left \{\|\xbar-\mu\|_2^2 \geq \frac{x}{2} \right \}
  +
  \PP \left \{\norm{\frac{1}{n} \sum_{i=1}^n
    (\xx_i - \mu)(\xx_i-\mu)^\top - \Sigma} \geq \frac{x}{2} \right \}
\end{equation*}
The first term is simply bounded by $5^{m} \exp \left \{- n x \rho/2
\right\}$ via an application of
equation~\eqref{eq:subgaussian-spectral-bound}.

We next bound the second term. By (\ref{eq:subgaussian}), there exists
a \(\rho' > 0\) such that:
\begin{equation*}
    \PP\left\{
        \vv^\top(\xx_i - \EE\xx_i)(\xx_i - \EE\xx_i)^\top\vv > x
    \right\}
    \leq
    \exp\left\{
        -\frac{x\rho'}{2}
    \right\}
\end{equation*}
It follows that
\(\EE\exp \left (t\vv^\top(\xx_i - \EE\xx_i)(\xx_i - \EE\xx_i)^\top\vv
  \right ) < \infty\)
for all \(t < \frac{\rho'}{2}\) and \(\norm{\vv} = 1\).  Then, there exists
a \(\tilde\rho\) such that:
\begin{equation*}
    \PP\left\{
        \left|
            \frac{1}{n}\sum_{i=1}^n
            \vv^\top
            [(\xx_i - \EE\xx_i)(\xx_i - \EE\xx_i)^\top - \Sigma]
            \vv
        \right|
        > \frac{x}{2}
    \right\}
    \leq
    \exp\left\{
        -\frac{n x^2 \tilde\rho}{4}
    \right\}
\end{equation*}
for all \(0 < x < \tilde\rho\) and \(\norm{\vv} = 1\). Thus
(\ref{eq:subgaus-sigma-spectral-bound}) follows immediately from
(\ref{eq:subgaussian-spectral-bound}) and the above bound on the
$\PP \left \{ \|\xbar-\mu\|_2^2 \geq \frac{x}{2} \right\}$.
\qed